%% file: main.tex
\documentclass{article}
\PassOptionsToPackage{table}{xcolor}
\PassOptionsToPackage{hyphens}{url}
\usepackage{iclr2027_conference,times}
\iclrfinalcopy
\usepackage[utf8]{inputenc}
\usepackage{graphicx}
\usepackage{hyperref}
\usepackage{url}
\usepackage{microtype}
\usepackage{booktabs}
\usepackage{multirow}
\usepackage{amsmath}
\usepackage{amsfonts}
\usepackage{amssymb}
\usepackage{enumitem}
\usepackage{xspace}
\usepackage{array}
\usepackage{makecell}
\usepackage{wrapfig}
\newcolumntype{L}[1]{>{\raggedright\arraybackslash}m{#1}}
\newcolumntype{C}[1]{>{\centering\arraybackslash}m{#1}}
\AddToHook{env/table/begin}{\setlength{\belowcaptionskip}{12pt}}
\AddToHook{env/table*/begin}{\setlength{\belowcaptionskip}{12pt}}
\usepackage{textcomp}
\usepackage[skins,breakable]{tcolorbox}

\newcommand{\mathds}[1]{\mathbb{#1}}

\newtcolorbox{examplebox}[1]{
  colback=gray!4, colframe=gray!55, coltitle=black,
  fonttitle=\bfseries\small, title={#1},
  boxrule=0.4pt, arc=2pt, left=5pt, right=5pt, top=4pt, bottom=4pt,
  enhanced
}
\newcommand{\exrole}[1]{\smallskip\noindent\textsc{\scriptsize #1}\\[-1pt]}
\newcommand{\exout}[1]{\smallskip\noindent{\scriptsize\textbf{#1}}\\[-1pt]}

\definecolor{langlightblue}{rgb}{0.3, 0.65, 1}
\definecolor{langblue}{rgb}{0, 0.4, 0.8}
\definecolor{langmildblue}{rgb}{0.0, 0.45, 0.73}
\definecolor{langdarkblue}{rgb}{0.0, 0.0, 0.61}
\definecolor{langred}{rgb}{0.81, 0.09, 0.13}
\definecolor{langlightgreen}{rgb}{0.80, 0.97, 0.85}
\definecolor{figureblue}{HTML}{315A83}
\definecolor{figurelightblue}{HTML}{DCEAF4}
\definecolor{langgreen}{rgb}{0.18, 0.55, 0.34}
\definecolor{langdarkgreen}{rgb}{0.0, 0.45, 0.38}
\definecolor{bingpink}{rgb}{1.0, 0.41, 0.71}
\definecolor{MidnightBlue}{RGB}{37,101,147}
\hypersetup{
  pdftitle={MedGuideX: Internalizing Decision Logic from Executable Guidelines into Large Language Models for Clinical Reasoning},
  pdfauthor={Yuhao Shen, Lang Cao, Simo Du, Yuqing Wang, Juexiao Zhou, Hao Peng, Yue Guo},
  colorlinks=true,
  citecolor=MidnightBlue,
  linkcolor=MidnightBlue,
  urlcolor=MidnightBlue
}
\newcommand{\gain}[1]{\textsubscript{\textbf{\textit{\textcolor{figureblue}{#1}}}}}

\graphicspath{{figures/}{./}}

\newcommand{\model}{\textit{\text{MedGuideX}}\xspace}

\title{MedGuideX: Internalizing Decision Logic from Executable Guidelines into Large Language Models for Clinical Reasoning}

\author{%
    \parbox[t]{\textwidth}{\raggedright
        \bfseries
        Yuhao Shen\textsuperscript{1,2,*},
        Lang Cao\textsuperscript{1,*},
        Simo Du\textsuperscript{3},
        Yuqing Wang\textsuperscript{3},
        Juexiao Zhou\textsuperscript{2},
        Hao Peng\textsuperscript{1},
        Yue Guo\textsuperscript{1}\\[4pt]
        \normalfont\small
        \mbox{\textsuperscript{1}\,University of Illinois Urbana-Champaign}\quad
        \mbox{\textsuperscript{2}\,The Chinese University of Hong Kong, Shenzhen}\quad
        \mbox{\textsuperscript{3}\,Albert Einstein College of Medicine}\quad
        \mbox{\textsuperscript{*}\,Equal contribution.}
    }%
}

\begin{document}
\vspace*{-15pt}
{\setlength{\tabcolsep}{0pt}\maketitle}
\vspace{-10pt}

\begin{abstract}

Clinical practice guidelines (CPGs) encode evidence-based decision logic that clinicians apply by evaluating patient variables, conditional criteria, and recommendation rules. However, existing methods often use CPGs as free-text training data or retrieval sources, underutilizing their procedural decision structure.
To better exploit this structure, we introduce a guideline-derived training pipeline that transforms CPG recommendations into executable clinical decision logic and uses it to generate factual and counterfactual question-answering data. These data teach models both guideline-supported decisions and how decisions change under different patient conditions.
Post-training a medical LLM on the generated data yields \model. Across four clinical reasoning benchmarks, \model achieves a 10.42\% relative improvement in average accuracy. 
Physician evaluation further shows that \model 
better recovers clinician-authored reasoning steps and produces physician-preferred rationales in faithfulness, validity, completeness, and conciseness.
Overall, our results show that executable decision logic from CPGs can be transformed into scalable supervision for building reliable medical LLMs.
\end{abstract}

\section{Introduction}

Large language models (LLMs) \citep{singh2025openai,yang2025qwen3} have shown strong potential in the medical domain, including electronic health records understanding, clinical case reasoning, and medical decision support \citep{cao2026ehr,wu2025medcasereasoning,lai2025doctor}. 
However, reliable clinical reasoning remains challenging. It requires models to integrate heterogeneous patient evidence, apply domain knowledge, compare plausible clinical decisions, handle uncertainty, and follow evidence-based decision logic \citep{bowen2006educational, nendaz2012diagnostic, sox2024medical}. 

Clinical practice guidelines (CPGs) offer a natural source of such decision logic. In clinical practice, clinicians apply guidelines by identifying patient variables, evaluating conditional criteria, and following recommendation rules. Thus, beyond textual medical knowledge, CPGs encode procedural decision structures for diagnosis, treatment, and disease management.

To better exploit this structure, we propose a guideline-grounded training pipeline for building LLMs with stronger clinical reasoning ability. We first collect high-quality, publicly available CPGs and transform their recommendations into executable functions that represent structured clinical decision logic.
Each function operates over patient variables and produces guideline-consistent decisions, enabling controlled data generation and automatic verification.
Based on these functions, we generate factual and counterfactual question-answering instances. Factual instances teach models guideline-supported decisions, while counterfactual instances teach how decisions should change when key patient conditions are modified.

Using this pipeline, we train \model, a medical LLM designed to internalize guideline-grounded clinical decision logic. Specifically, we post-train the base model with supervised fine-tuning (SFT) and reinforcement learning (RL) on the generated factual and counterfactual data. Experiments on four clinical reasoning benchmarks show that \model substantially improves over its base model and achieves strong performance among open-source medical LLMs. 
Compared with Qwen3.5-9B~\citep{qwen2026qwen35}, \model-9B achieves relative improvements of $26.64\%$, $9.45\%$, $4.41\%$, and $11.67\%$ on MedCaseReasoning~\citep{wu2025medcasereasoning}, MedQA~\citep{jin2021disease}, MIMIC-CDM-FI~\citep{hager2024evaluation}, and ER-Reason~\citep{mehandru2025er}, respectively.
Notably, the larger relative gains appear on the lower-accuracy benchmarks, MedCaseReasoning and ER-Reason, suggesting that guideline-derived supervision is particularly useful for challenging reasoning settings. 
Physician evaluation further shows that \model better recovers clinician-authored reasoning steps and produces physician-preferred rationales in faithfulness, validity, completeness, and overall quality.

In summary, our contributions are:
\begin{itemize}[leftmargin=*, itemsep=0pt, labelsep=5pt, topsep=0pt]
    \item We propose a guideline-derived post-training pipeline that transforms CPGs into executable clinical decision logic and uses it to generate factual and counterfactual QA supervision.

    \item We train \model, a medical LLM that internalizes guideline-grounded clinical decision logic through SFT and RL.
    
    \item We conduct experiments on four medical reasoning benchmarks, showing that \model improves over the base model and similarly sized medical LLMs, while producing higher-quality clinical rationales.
    
\end{itemize}

\begin{figure}[t]
    \centering
    \includegraphics[width=\textwidth]{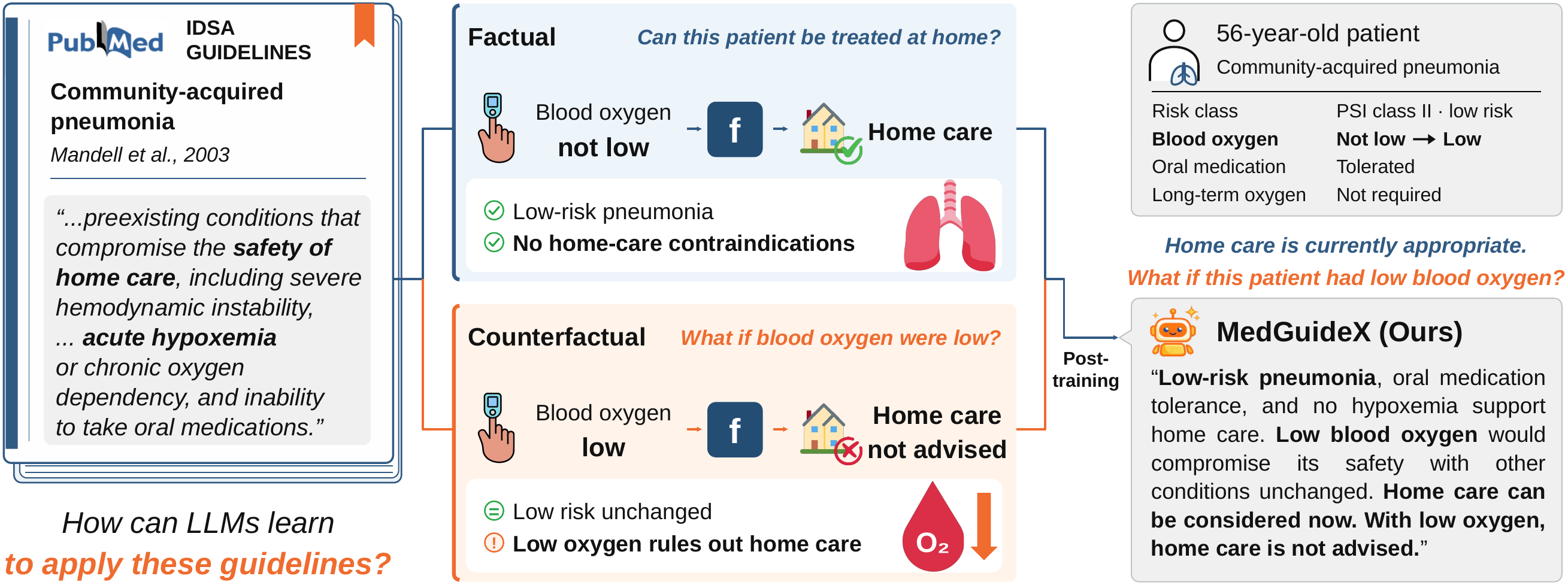}
    \caption{\textbf{Overview of MedGuideX.} We convert clinical practice guidelines into executable decision rules and use them to generate factual and counterfactual supervision. Post-training with this supervision improves clinical reasoning and helps the model apply guideline rules to patient cases and adjust recommendations when relevant patient conditions change.}
    \nocite{mandell2003pneumonia}
    \label{fig:overview}
\end{figure}

\section{Related Work}
\label{sec:related_work}

\begin{figure}[t]
    \centering
    \includegraphics[width=\textwidth]{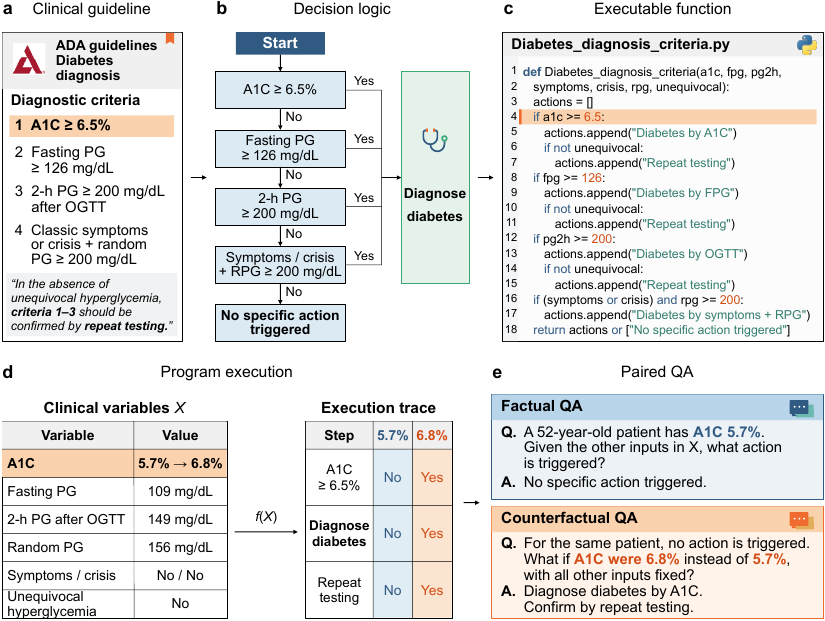}
    \caption{\textbf{From clinical guidelines to executable supervision.} \textbf{a.} Guideline recommendations~\citep{colberg2010exercise} define conditions and actions. \textbf{b.} Recommendations are organized into decision logic. \textbf{c.} A deterministic converter produces executable functions. \textbf{d.} Execution on factual and intervened patient states produces action labels and traces. \textbf{e.} These outputs provide factual and counterfactual QA supervision, tying each training target to the conditions that produce it.}
    \label{fig:pipeline}
\end{figure}

\subsection{LLMs for Clinical Reasoning}
Medical LLMs are trained on large-scale medical corpora, clinical notes, and case reports~\citep{chen2023meditron,han2023medalpaca,labrak2024biomistral,garcia2025aloe,wu2025medcasereasoning}, and enhanced through retrieval, multi-agent diagnosis, or clinical-feedback post-training~\citep{cao2026ehr,li2025macd,chen2024huatuogpt,lai2025doctor}. These heterogeneous sources broaden medical knowledge and case-level reasoning, but often provide implicit, noisy, or incomplete supervision weakly aligned with explicit clinical decision procedures~\citep{gu2025clinical,li2026joint,yoo2025model,yang2024towards}.

\subsection{Clinical Practice Guidelines for LLMs}
CPG-based methods retrieve guideline passages for prompting~\citep{schubert2025guideline,deng2026cpgprompt,oniani2024enhancing,li2023meddm}, construct guideline-grounded SFT rationales or train directly on guideline text~\citep{staniek2025training,chen2023meditron}, and use guideline constraints as RL rewards~\citep{tziakouri2025reinforcement,gu2025clinical}. Prior work uses clinical decision structures at inference and instantiates guideline rules for supervised training. Our framework compiles guideline recommendations into functions that generate factual and interventional supervision for staged clinical post-training.

\subsection{Post-Training and Executable Counterfactual Reasoning}
Intermediate reasoning steps such as chain-of-thought can elicit complex reasoning~\citep{wei2022emergent,suzgun2023challenging,wei2022chain,cao2024graphreason}, while post-training seeks to internalize such behavior. SFT uses filtered self-generated solutions~\citep{yuan2023scaling,tong2024dart} or rationales distilled from stronger models~\citep{muennighoff2025s1}, whereas RL~\citep{kaelbling1996reinforcement} optimizes task-level objectives beyond next-token prediction~\citep{lightman2024let,uesato2022solving,guo2025deepseek}. Large-scale RL and its extensions to search and structured data further demonstrate this paradigm~\citep{guo2025deepseek,jiang2025deepretrieval,jin2025search,cao2025fortune}. Counterfactual reasoning can also strengthen model reasoning and reveal failures missed by standard QA~\citep{chen2025counterbench,you2026improving,vashishtha2025executable}. Our work combines these directions by deriving verifiable factual and counterfactual clinical supervision from executable, evidence-based guidelines.

\section{Methodology}
\label{sec:method}

\subsection{Logic Formulation}
We define a guideline recommendation as an actionable clinical decision rule extracted from a complete CPG. A recommendation specifies how a clinical decision should be made under certain patient conditions, such as diagnosis, treatment, screening, or disease management. For example, a recommendation may state that a patient with severe symptoms should be referred for further evaluation.

We formalize each guideline recommendation as a function
\begin{equation}
    Y = f(X), \qquad f : \mathcal{X} \to \mathcal{Y},
\end{equation}
where $X = (x_1, \dots, x_n) \in \mathcal{X}$ denotes a vector of clinical variables, $f$ denotes the conditional decision logic encoded by the recommendation, and $Y \in \mathcal{Y}$ is the guideline-prescribed output. We instantiate $f$ as a finite decision tree $T_f$, whose internal nodes are atomic predicates $a \in \mathcal{A}_f$ over $X$ and whose leaves are outputs in $\mathcal{Y}$. Each predicate takes the form $a_i = \mathbf{1}[g_i(X)]$, where $g_i$ is a guideline-defined condition, such as $\textit{age} \ge 65$ or $\textit{eGFR} < 30$.

Given an input $X$, executing $T_f$ activates a path
\begin{equation}
    \pi_f(X) = (a_{i_1}, a_{i_2}, \dots, a_{i_k}),
\end{equation}
whose predicates jointly select the leaf output $Y=f(X)$. This path is the verifiable unit of guideline logic that we aim for the model to internalize: it should not only predict the correct output, but also reason through a path consistent with $\pi_f(X)$.

We next describe how we collect CPGs and construct factual and counterfactual QA data from them (\S~\ref{sec:data}), followed by how we train \model using SFT and RL (\S~\ref{sec:training}). Figure~\ref{fig:pipeline} illustrates the full pipeline.

\subsection{Data Construction}
\label{sec:data}

The central artifact of data preparation is an executable implementation of $f$, whose control flow mirrors the decision tree $T_f$. This executable form provides deterministic targets for factual and counterfactual supervision. Factual GRPO checks the final action against its execution target, while the released counterfactual verifier also checks the inferred hidden state by re-execution. In this formulation, the inputs correspond to structured patient information and clinical scenarios, while the executable function $f$ represents structured clinical reasoning grounded in guideline decision logic.

\subsubsection{Guideline Curation}
\input{tables/data_stats}
Our initial guideline source is an open CPG collection based on the corpus used to train MEDITRON~\citep{chen2023meditron}. However, many documents in this collection are noisy, low quality, or near-duplicate guidelines, which can substantially degrade the quality of downstream QA data. We therefore apply a curation pipeline to retain high-quality guidelines.

We first restrict the corpus to guidelines from U.S.-based sources, since clinical recommendations may vary across countries, healthcare systems, and organizations. Specifically, we retain guidelines from the Centers for Disease Control and Prevention (CDC)\footnote{\url{https://www.cdc.gov/}} and PubMed\footnote{\url{https://pubmed.ncbi.nlm.nih.gov/}}. We then use an LLM to extract structured metadata for each guideline, including the disease or drug, target age group, race, gender, and publication date. Guidelines with identical metadata are treated as duplicates, and only the most recent version is retained. In addition, we instruct the LLM to discard incomplete documents directly. After curation, we obtain a filtered subset of CPGs.

\subsubsection{Executable Transformation}

\paragraph{Recommendation Extraction.}
We first split each document into recommendation-oriented chunks, where each chunk contains one or more complete guideline recommendations. This produces a set of guideline recommendation passages. An LLM extractor then identifies recommendation candidates from each chunk. For each candidate, we ask the extractor to identify the target population, clinical condition, recommended action, relevant exceptions, and evidence grade, when available. We then validate these candidates and retain only usable recommendations. Specifically, we discard candidates that do not describe a concrete clinical action or cannot be expressed as a condition-action rule. We also remove near-duplicates, where multiple recommendations describe highly similar populations, conditions, and actions.

\paragraph{Decision-Tree Validation.}
Each retained recommendation is converted into a decision tree $T_f$, which specifies the required input variables, decision conditions, and final outputs. An LLM validator then checks whether the tree is complete, whether each condition is clear, whether every branch leads to an output, and whether all variables are supported by the source guideline.

\paragraph{Compilation to Executable Function.}
Each validated tree is compiled into an executable Python function that takes the variables $X$ as input and returns the guideline output $f(X)$. We further check whether the function is syntactically correct, executable on sampled inputs, and consistent with the original decision tree.

\subsubsection{QA Synthesis}

\paragraph{Factual QA Synthesis.}
For each executable function $f$, we sample complete clinical variable assignments $X$ and execute $f$ to obtain the guideline output $Y_{\text{obs}} = f(X)$. A naive sampling strategy would produce too many easy or default cases, such as no-action recommendations. To avoid this imbalance, we enforce two constraints: (1) \textbf{\textit{Path coverage}}: the generated data should cover all decision conditions in the tree, and (2) \textbf{\textit{Output balance}}: no-diagnosis outputs, meaning that no diagnosis is recommended for the current inputs, should not dominate the dataset.

After applying the coverage and balancing constraints, we obtain the factual QA set. For each sample, we use an LLM to generate a step-by-step reasoning trace from the underlying Python function $f$, input variables $X$, and executed output $Y_{\text{obs}}$. The reasoning trace verbalizes the executed decision path $\pi_f(X)$ and is stored alongside the QA pair for SFT training.

\input{tables/main_results}
\begin{figure}[t]
    \centering
    \includegraphics[width=\textwidth]{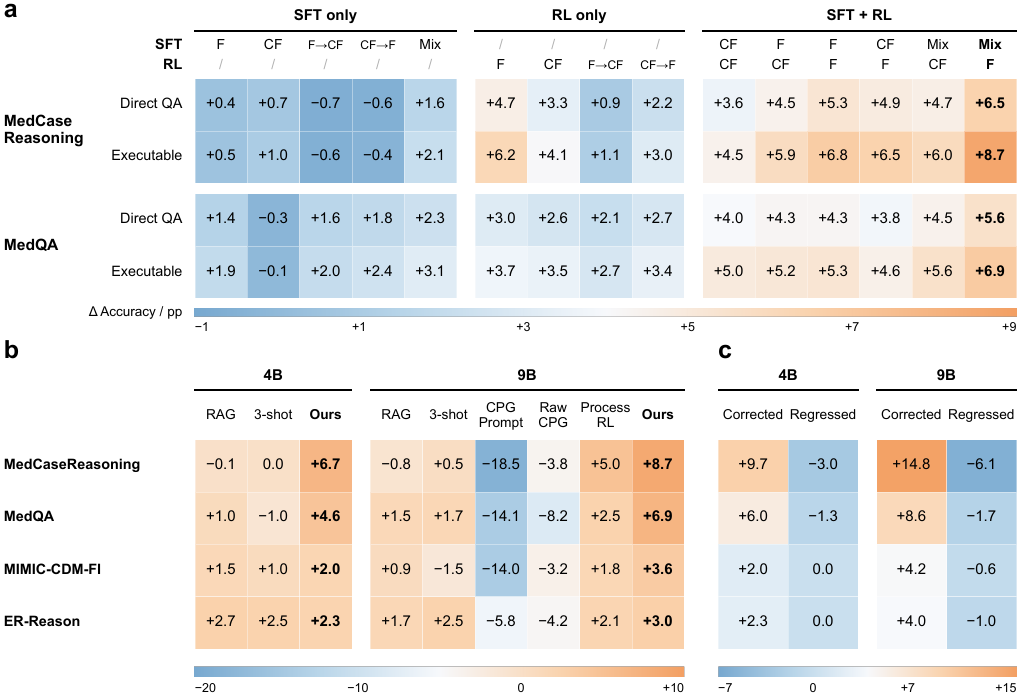}
    \caption{\textbf{Training ablations, guideline-use baselines, and answer transitions.} \textbf{a.} Accuracy gains over Qwen3.5-9B for Direct QA and executable supervision across 15 training configurations. F denotes factual data, CF counterfactual data, and Mix an equal mixture. Arrows indicate sequential training within a stage; grey slashes mark unused stages. \textbf{b.} Gains over the corresponding 4B or 9B backbone for alternative guideline uses and \model, with full accuracy results in Appendix Table~\ref{tab:guideline_baselines}. \textbf{c.} Wrong-to-right and right-to-wrong answer transitions, divided by all test items. Regressions are plotted as negative values, so their sum with corrections gives the net accuracy gain. Gains are in percentage points, with a separate color scale for each panel.}
    \label{fig:ablation}
\end{figure}
\input{tables/guideline_decisions}
\input{tables/reasoning_anlaysis}

\paragraph{Counterfactual QA Synthesis.}
We further generate counterfactual QA data to train the model to reason about hypothetical changes in patient conditions. For each counterfactual example, we first sample a complete variable assignment $X$ over the inputs of $f$ and execute $f$ to obtain the factual outcome $Y_{\text{obs}} = f(X)$. We then partition the variables into three disjoint parts:
\begin{itemize}[leftmargin=*, itemsep=0pt, labelsep=5pt, topsep=0pt]
    \item $X_{\text{obs}}$: observed variables that are shown and remain unchanged.
    \item $X_{\text{hid}}$: hidden variables that are not shown.
    \item $x_{\text{int}}$: a single observed variable modified by the intervention, meaning that its value is changed while all other observed variables are held fixed.
\end{itemize}

We write $\hat{x}_{\text{int}}$ for the value of $x_{\text{int}}$ after intervention. Only $x_{\text{int}}$ changes, while $X_{\text{obs}}$ and $X_{\text{hid}}$ are held fixed. The factual and counterfactual outcomes can be written as
\begin{equation}
\label{eq:cf_outcomes}
\begin{aligned}
    Y_{\text{obs}} &= f\big(X_{\text{obs}}, X_{\text{hid}}, x_{\text{int}}\big),\\
    Y_{\text{cf}}^{\star} &= f\big(X_{\text{obs}}, X_{\text{hid}}, \hat{x}_{\text{int}}\big).
\end{aligned}
\end{equation}
By construction, $Y_{\text{obs}}$ and $Y_{\text{cf}}^{\star}$ differ only through the intervention on $x_{\text{int}}$.

The model receives $X_{\text{obs}}$, the identity of the hidden variables, the original and intervened values $x_{\text{int}}$ and $\hat{x}_{\text{int}}$, and the factual outcome $Y_{\text{obs}}$. It must then predict $Y_{\text{cf}}^{\star}$ through three steps:
\begin{enumerate}[leftmargin=*, itemsep=0pt, labelsep=5pt, topsep=0pt]
    \item \textbf{Abduction.} Infer hidden values $\hat{X}_{\text{hid}}$ that are consistent with the observed context and reproduce $Y_{\text{obs}}$ when executed through $f$.
    \item \textbf{Intervention.} Replace $x_{\text{int}}$ with $\hat{x}_{\text{int}}$ while keeping $X_{\text{obs}}$ and $\hat{X}_{\text{hid}}$ fixed.
    \item \textbf{Prediction.} Execute $f$ on the resulting complete input to obtain the predicted counterfactual outcome $\hat{Y}_{\text{cf}}$.
\end{enumerate}

At the data level, we execute $f$ on the factual input, apply the intervention, and execute $f$ again to generate interventional scenarios. We retain only cases where the intervention changes the outcome and discard those with unchanged outputs. Each counterfactual sample is paired with a reasoning trace generated by an LLM, which verbalizes the three steps of abduction, intervention, and prediction over the executable function $f$. Overall data pipeline statistics are shown in Appendix~\ref{sec:data_details}.

\subsection{Post-Training}
\label{sec:training}

We post-train \model using prepared factual and counterfactual QA sets.

\paragraph{Optimization objectives.}
For each guideline recommendation, we obtain an executable function $f$. Given a complete patient scenario $X$, the function returns the guideline output
\begin{equation}
\label{eq:factual_label}
    Y = f(X).
\end{equation}

A factual QA instance asks the model to determine the guideline output for the complete patient scenario $X$. For factual examples, each oracle trajectory contains the patient scenario, the executed guideline output, and a rationale that explains why the scenario leads to that output $Y$.

For counterfactual QA examples, each trajectory follows an abduction--intervention--prediction structure. The prompt provides $X_{\text{obs}}$, the original value $x_{\text{int}}$, the intervened value $\hat{x}_{\text{int}}$, and the factual guideline output $Y_{\text{obs}}$; the hidden variable is not shown to the model. The model must first infer a hidden value that makes the factual observation consistent with the executable guideline, then apply the intervention while keeping the inferred hidden value fixed, and finally predict the counterfactual guideline output $\hat{Y}_{\text{cf}}$.

The optimization objective is to train the model so that its predicted guideline output $\hat{Y}$ matches the oracle output $Y$ as closely as possible in both factual and counterfactual QA, thereby improving its ability to perform clinical reasoning aligned with the CPGs from which the synthetic data are derived.

\paragraph{SFT.}
The SFT stage trains the model to imitate oracle trajectories synthesized from the executable guideline functions. This stage serves as a cold start for subsequent RL training by injecting guideline-grounded clinical knowledge and exposing the model to a structured clinical reasoning framework.

\paragraph{RL.}
After SFT, we apply GRPO~\citep{shao2024deepseekmath} to optimize the model on its own sampled responses. For a sampled response $o$, we define a format reward $r_{\text{fmt}}(o)$ and an answer reward $r_{\text{answer}}(o)$. The format reward checks whether $o$ satisfies the required response format, while the answer reward checks whether the parsed answer is correct. The final reward is
\begin{equation}
\label{eq:rl_reward}
r(o) =
\begin{cases}
-1, & r_{\text{fmt}}(o) = -1, \\
r_{\text{answer}}(o), & r_{\text{fmt}}(o) = 0.
\end{cases}
\end{equation}
Here, $r_{\text{fmt}}(o)=-1$ indicates an invalid response format, $r_{\text{fmt}}(o)=0$ indicates a valid format, and $r_{\text{answer}}(o)\in\{0,1\}$ indicates whether the task-specific correctness check passes.

For factual prompts, the reward compares the parsed final answer $\hat{Y}(o)$ with the execution target $Y^{\star}=f(X)$. The factual answer reward is
\begin{equation}
\label{eq:factual_correctness}
    r_{\text{F}}(o)
    =
    \mathds{1}\!\left[\hat{Y}(o) \equiv Y^{\star}\right].
\end{equation}

For counterfactual prompts, let $\hat{X}_{\text{hid}}(o)$ and $\hat{Y}_{\text{cf}}(o)$ denote the inferred hidden value and predicted answer. Correctness requires recovering the sampled hidden state $X_{\text{hid}}^{\star}$, reproducing $Y_{\text{obs}}$ under the original factual context, and matching the executed counterfactual label.
\begin{equation}
\label{eq:counterfactual_original_correctness}
\begin{aligned}
    r_{\text{CF}}(o)
    =
    &\ \mathds{1}\!\left[\hat{X}_{\text{hid}}(o) = X_{\text{hid}}^{\star}\right] \\
    &\cdot
    \mathds{1}\!\left[f\big(X_{\text{obs}}, \hat{X}_{\text{hid}}(o), x_{\text{int}}\big) = Y_{\text{obs}}\right] \\
    &\cdot
    \mathds{1}\!\left[\hat{Y}_{\text{cf}}(o) = Y_{\text{cf}}^{\star}\right].
\end{aligned}
\end{equation}
Thus, the counterfactual reward requires the model to recover the intended hidden value and predict the correct counterfactual output.

\paragraph{Training procedure.}
We perform SFT on the balanced factual and counterfactual QA mixture, then initialize GRPO from the SFT checkpoint. The reward implementation supports both factual and counterfactual QA. Our final \model configuration uses mixed factual and counterfactual SFT followed by factual GRPO, as selected by the ablation study in Appendix~\ref{sec:ablation}. Counterfactual GRPO is evaluated as an alternative configuration in the same ablation.

\section{Experiments}
\label{sec:exp}

\subsection{Experimental Setup}
\label{sec:setup}
We evaluate \model on MedQA~\citep{jin2021disease}, MedCaseReasoning~\citep{wu2025medcasereasoning}, MIMIC-CDM-FI~\citep{hager2024evaluation}, and ER-Reason~\citep{mehandru2025er}. Benchmark training sets are excluded from post-training. We compare frontier and medical LLMs, Qwen3.5-4B and 9B, and alternative guideline-based methods. We parse MedQA answers deterministically. On the other three benchmarks, GPT-5.4 mini~\citep{openai2026gpt54mini} with \texttt{xhigh} reasoning effort judges diagnostic equivalence. Qwen3.5 and \model use matched decoding. Appendices~\ref{sec:benchmark_details}, \ref{sec:training_details}, and~\ref{sec:guideline_baselines} give the protocols.

\subsection{Main Results}
\label{sec:main_results}
Guideline post-training improves every benchmark at both scales and reverses the ranking between a smaller adapted model and a larger backbone. In Table~\ref{tab:main_results}, \model-4B exceeds Qwen3.5-9B on all four benchmarks, with average accuracy of $55.18\%$ against $53.29\%$. Its $3.9$-point improvement exceeds the $2.0$-point gain from scaling the unadapted backbone from 4B to 9B. At 9B, post-training raises average accuracy to $58.85\%$, with gains of $8.7$ points on MedCaseReasoning, $6.9$ on MedQA, $3.6$ on MIMIC-CDM-FI, and $3.0$ on ER-Reason. Guideline supervision therefore complements model scale, with larger gains at 9B on every benchmark.

Both \model scales lead the listed medical LLMs in average accuracy. \model-9B exceeds Hulu-Med-7B by $7.6$ points and achieves $41.36\%$ on MedCaseReasoning against the strongest medical comparator at $31.22\%$. Its average deficit to GPT-5.0 narrows from the backbone's $12.0$ points to $6.5$, while MIMIC-CDM-FI accuracy reaches $85.20\%$ against $85.10\%$.

Accuracy gains combine error correction with uneven retention of previously solved cases. Figure~\ref{fig:ablation}c shows $133$ corrections against $55$ regressions on MedCaseReasoning at 9B, compared with $110$ against $22$ on MedQA. The correction-to-regression ratios are $2.4$ and $5.0$, respectively. MedQA thus preserves solved cases more effectively despite its smaller net gain. Corrections exceed regressions in all eight benchmark and scale comparisons. Appendix~\ref{sec:transition_details} gives the complete counts.

\subsection{Comparison of Guideline-Based Methods}
\label{sec:direct_control}
Guideline access alone yields limited transfer in Figure~\ref{fig:ablation}b. Retrieval and three-shot prompting each improve 9B average accuracy by $0.8$ points, compared with $5.6$ for \model. Raw CPG fine-tuning and our CPGPrompt adaptation~\citep{deng2026cpgprompt} reduce accuracy by $4.8$ and $13.1$ points.

Execution adds value after patient questions and training resources are fixed. Direct QA generates answers and rationales from guideline text; executable supervision obtains labels from $f$ and rationales from its trace. Both arms use the same Qwen3.5-9B student inputs, including the function, matched questions, data volume, training budgets, and GRPO scoring rules. With mixed SFT followed by factual RL, executable supervision exceeds Direct QA by $2.2$ points on MedCaseReasoning and $1.3$ on MedQA in Figure~\ref{fig:ablation}a. Appendix~\ref{sec:direct_qa_protocol} specifies the comparison.

Additional rationale scoring weakens factual RL. From the same mixed SFT checkpoint, adding an LLM guideline-consistency reward lowers average accuracy from $58.85\%$ to $56.15\%$, with declines on all four benchmarks in Figure~\ref{fig:ablation}b. The executed action supplies the stronger reward objective in this comparison. Appendix~\ref{sec:baseline_process_reward} details the shared initialization and reward construction.

\subsection{Training Strategy Ablations}
\label{sec:training_ablations}
Counterfactual supervision contributes most through joint exposure during SFT. In Figure~\ref{fig:ablation}a, mixed executable SFT yields gains of $2.1$ points on MedCaseReasoning and $3.1$ on MedQA, outperforming factual-only, counterfactual-only, and sequential SFT. Both sequential orders fall below the backbone on MedCaseReasoning, showing that exposure to both data types alone does not reproduce the benefit of mixing them.

Factual RL provides the stronger optimization stage. Starting from the backbone, it gains $6.2$ and $3.7$ points on the two benchmarks, compared with $4.1$ and $3.5$ for counterfactual RL. Initializing factual RL from mixed SFT adds a further $2.5$ and $3.2$ points. At that checkpoint, replacing factual RL with counterfactual RL lowers the final gains from $8.7$ and $6.9$ to $6.0$ and $5.6$. Counterfactual data are most effective in preparing the model for factual optimization.

The same schedule maximizes both supervision arms, while executable targets outperform Direct QA in all $15$ matched configurations on both benchmarks. The execution advantage therefore persists across training schedules. Appendices~\ref{sec:ablation_schedules} and~\ref{sec:direct_qa_protocol} specify the schedules and scoring.

\subsection{Decision Behavior on Held-Out Guidelines}
\label{sec:decision_behavior}
Following the behavioral testing approach of \citet{ribeiro2020checklist}, we vary one patient finding at a time to test when a model revises a guideline recommendation and when it retains it. We draw guidelines from the same EPFL MEDITRON collection as the training corpus~\citep{chen2023meditron} and exclude the evaluation documents from both SFT and RL. GPT-6.1 Sol~\citep{openai2026gpt61sol} at \texttt{xhigh} generates $100$ original cases, two edits per case, and an action pseudolabel for each version. One edit preserves the recommended action and the other changes it. The backbone and the two models compared in Section~\ref{sec:direct_control} receive identical guideline passages and complete patient information. Each answers the three versions in separate queries.

Across the $300$ cases in Table~\ref{tab:guideline_decisions}, action accuracy rises from $58.7\%$ for Qwen3.5-9B to $64.3\%$ for matched Direct QA and $68.3\%$ for \model-9B. The largest gain over Direct QA comes when the recommendation must change. On the $100$ changing edits, \model gives the correct new action in $61$ cases versus $54$ for Direct QA. Requiring the original action to be correct as well yields $51$ successful pairs versus $41$. On the $100$ preserving edits, \model gets both actions right in $59$ pairs versus $52$. The counterfactual training filter in Section~\ref{sec:data} retains only edits that change the action, yet the gain extends to recognizing when the original recommendation still applies.

A correct recommendation can still rest on the wrong branch. For a written path to earn credit, the response must identify the decisive patient facts, apply the relevant guideline rule and any exception, and derive the correct action. GPT-5.4 mini~\citep{openai2026gpt54mini} and GPT-6.1 Sol each score the same $300$ answers per model in three runs at \texttt{xhigh}. The first judge credits $58.9\%$ of \model paths versus $55.3\%$ for Direct QA, and the second credits $60.1\%$ versus $55.7\%$. The gain thus extends to the stated basis for the recommendation, though both path scores fall short of \model's $68.3\%$ action accuracy. Appendix~\ref{sec:heldout_eval_protocol} gives the scoring protocol and Appendix~\ref{sec:heldout_prompts} gives the prompts.

\subsection{Clinical Rationale Evaluation}
\label{sec:reasoning_eval}
Clinical coverage distinguishes the rationales in Table~\ref{tab:reasoning_eval}. \model-9B covers $62.8\%$ of clinician-authored MedCaseReasoning reference steps against $50.8\%$ for GPT-5.0, a $12.0$-point gain. Appendix~\ref{sec:mcr_protocol} specifies the recall assessment.

Two physicians provide $40$ blinded annotations on $30$ cases correctly diagnosed by both models, including $10$ shared cases. Selection favored \model rationales, and response order is randomized. They assess \textbf{Evidence faithfulness}, \textbf{Reasoning validity}, \textbf{Reasoning completeness}, and \textbf{Reasoning conciseness}. Appendix~\ref{sec:human_study} details the study.

Within this sample, non-tie win rates reach $92.3\%$ for Reasoning completeness, $85.0\%$ for Evidence faithfulness, and $79.3\%$ for Reasoning validity, compared with $51.5\%$ for Reasoning conciseness. The pooled rate is $76.9\%$. Completeness receives $36$ preferences against $3$, while conciseness receives $17$ against $16$. Correct diagnoses thus coexist with a substantial difference in reasoning coverage, aligning the physician preference with the reference-step recall advantage.

\section{Conclusion}
We introduced \model, which converts clinical guideline recommendations into executable factual and counterfactual supervision for post-training. Mixed SFT followed by factual GRPO improves four clinical reasoning benchmarks at two model scales. The direct QA comparison shows additional gains from execution-derived supervision, and the held-out guideline evaluation shows higher action accuracy on both decision-preserving and decision-changing edits. Executable guidelines provide a practical source of supervision that ties clinical decisions to explicit patient conditions.

\clearpage
\bibliography{references}
\bibliographystyle{iclr2027_conference}

\clearpage
\appendix
\input{appendix}

\end{document}

%% file: tables/data_stats.tex
\begin{wraptable}{r}{0.50\textwidth}
\vspace{-\intextsep}
\setlength{\abovecaptionskip}{0pt}
\setlength{\belowcaptionskip}{12pt}
\caption{Corpus statistics from guideline curation to QA synthesis. Recommendation and program counts follow automated filtering and validation. QA counts describe the generated supervision pool.}
\label{tab:data_stats}
\centering
\fontsize{9}{10}\selectfont
\setlength{\tabcolsep}{5pt}
\renewcommand{\arraystretch}{1.1}
\begin{tabular}{@{}L{0.79\linewidth}C{\dimexpr0.21\linewidth-2\tabcolsep\relax}@{}}
\toprule
\multicolumn{1}{@{}l}{\textbf{Pipeline stage}} & \textbf{Count} \\
\midrule
Curated guideline documents & $841$ \\
\quad CDC & $607$ \\
\quad PubMed & $234$ \\
Guideline chunks & $3{,}127$ \\
Recommendation candidates & $4{,}750$ \\
Validated actionable recommendations & $3{,}796$ \\
Validated decision trees & $2{,}800$ \\
Executable guideline functions & $\mathbf{2{,}793}$ \\
\midrule
Factual QA samples & $\mathbf{4{,}963}$ \\
Counterfactual QA candidates & $\mathbf{5{,}028}$ \\
\bottomrule
\end{tabular}
\end{wraptable}

%% file: tables/main_results.tex
\begin{table*}[t]
\caption{Accuracy in percent on four clinical reasoning benchmarks for frontier LLMs, medical LLMs, Qwen3.5 backbones, and \model. Bold and italics mark the best and second-best scores among medical LLMs and our models. Blue subscripts give relative percentage improvements over the corresponding backbone.}
\label{tab:main_results}
\centering
\fontsize{9}{10}\selectfont
\setlength{\tabcolsep}{1.5pt}
\renewcommand{\arraystretch}{1.2}
\begin{tabular}{@{}L{0.46\textwidth}*{5}{C{\dimexpr0.108\textwidth-2\tabcolsep\relax}}@{}}
\toprule
\multicolumn{1}{c}{\makecell[c]{\textbf{Model}}}
& \multicolumn{1}{c}{\makecell[c]{\textbf{MedCase}\\\textbf{Reasoning}}}
& \multicolumn{1}{c}{\makecell[c]{\textbf{MedQA}}}
& \multicolumn{1}{c}{\makecell[c]{\textbf{MIMIC-}\\\textbf{CDM-FI}}}
& \multicolumn{1}{c}{\makecell[c]{\textbf{ER-}\\\textbf{Reason}}}
& \multicolumn{1}{c}{\makecell[c]{\textbf{Average}}} \\
\midrule
\multicolumn{6}{@{}l@{}}{\textbf{\textit{Frontier LLM}}} \\
GPT-5.0~\citep{singh2025openai}                       & 54.96 & 87.43 & 85.10 & 33.70 & 65.30 \\
DeepSeek-V4~\citep{deepseek2026v4}                    & 52.51 & 88.92 & 90.10 & 24.00 & 63.88 \\
Kimi-K2.6~\citep{moonshot2026kimi26}                          & 23.63 & 79.26 & 74.60 & 34.40 & 52.97 \\
Claude-Haiku-4.5~\citep{anthropic2025claude-haiku-4.5} & 40.02 & 48.08 & 81.60 & 34.30 & 51.00 \\

\midrule
\multicolumn{6}{@{}l@{}}{\textbf{\textit{Open-source Medical LLM}}} \\
Clinical-R1-3B~\citep{gu2025clinical}          & 13.60 & 52.55 & 62.30 & 19.10 & 36.89 \\
MediPhi-Instruct-4B~\citep{corbeil2025modular} & 17.06 & 53.34 & 72.60 & 22.20 & 41.30 \\
MedGemma-1.5-4B~\citep{sellergren2025medgemma} & 16.39 & 67.79 & 80.10 & 20.80 & 46.27 \\
Hulu-Med-7B~\citep{jiang2025hulu}              & 31.22 & 72.98 & 78.70 & 22.20 & 51.27 \\
Lingshu-7B~\citep{xu2025lingshu}               & 23.86 & 64.49 & 78.00 & 26.20 & 48.14 \\
Llava-Med-v1.5-7B~\citep{li2023llava}          & 16.72 & 44.78 & 60.10 & 21.20 & 35.70 \\
MedAlpaca-7B~\citep{han2023medalpaca}          & 12.26 & 38.81 & 41.00 & 11.00 & 25.77 \\
BioMistral-7B~\citep{labrak2024biomistral}     & 17.28 & 45.80 & 35.60 & 24.80 & 30.87 \\
HuatuoGPT-o1-8B~\citep{chen2024huatuogpt}      & 19.29 & 71.09 & 68.10 & 22.00 & 45.12 \\
Llama-Aloe-Beta-8B~\citep{garcia2025aloe}   & 21.18 & 64.65 & 80.80 & \textbf{29.10} & 48.93 \\
MedReason-8B~\citep{wu2025medreason}           & 27.76 & 70.86 & 73.00 & 22.40 & 48.50 \\

\midrule
\multicolumn{6}{@{}l@{}}{\textbf{\textit{Ours}}} \\
Qwen3.5-4B~\citep{qwen2026qwen354b} & 28.65 & 72.43 & 80.40 & 23.60 & 51.27 \\
\rowcolor{figurelightblue}
\textbf{\model-4B}
  & \textit{35.34\gain{+23.3\%}}
  & \textit{77.06\gain{+6.4\%}}
  & \textit{82.40\gain{+2.5\%}}
  & 25.90\gain{+9.7\%}
  & \textit{55.18\gain{+7.6\%}} \\
\addlinespace[2pt]
Qwen3.5-9B~\citep{qwen2026qwen35} & 32.66 & 73.21 & 81.60 & 25.70 & 53.29 \\
\rowcolor{figurelightblue}
\textbf{\model-9B}
  & \textbf{41.36\gain{+26.6\%}}
  & \textbf{80.13\gain{+9.4\%}}
  & \textbf{85.20\gain{+4.4\%}}
  & \textit{28.70\gain{+11.7\%}}
  & \textbf{58.85\gain{+10.4\%}} \\
\bottomrule
\end{tabular}
\end{table*}

%% file: tables/guideline_decisions.tex
\begin{table}[t]
\centering
\caption{Decisions on held-out guidelines. Action accuracy uses $100$ cases per category and $300$ overall. Pair accuracy requires both actions to be correct. Written-path accuracy requires the stated guideline conditions to support the correct action. All scores are percentages.}
\label{tab:guideline_decisions}
\small
\setlength{\tabcolsep}{1.7pt}
\renewcommand{\arraystretch}{1.15}
\begin{tabular*}{\linewidth}{@{\extracolsep{\fill}}l*{8}{c}@{}}
\toprule
& \multicolumn{4}{c}{\textbf{Single action}}
& \multicolumn{2}{c}{\textbf{Paired actions}}
& \multicolumn{2}{c}{\textbf{Written path}} \\
\cmidrule(lr){2-5}\cmidrule(lr){6-7}\cmidrule(l){8-9}
\textbf{Model} & Original & Preserve & Change & Overall & Preserve & Change
& \makecell[c]{GPT-5.4\\mini} & \makecell[c]{GPT-6.1\\Sol} \\
\midrule
Qwen3.5-9B & 66.0 & 61.0 & 49.0 & 58.7 & 47.0 & 35.0 & 51.8 & 50.2 \\
Direct QA 9B & 72.0 & 67.0 & 54.0 & 64.3 & 52.0 & 41.0 & 55.3 & 55.7 \\
\model-9B & \textbf{74.0} & \textbf{70.0} & \textbf{61.0} & \textbf{68.3} & \textbf{59.0} & \textbf{51.0} & \textbf{58.9} & \textbf{60.1} \\
\midrule
$\Delta$ vs.\ Direct & +2.0 & +3.0 & +7.0 & +4.0 & +7.0 & +10.0 & +3.6 & +4.4 \\
\bottomrule
\end{tabular*}\par\smallskip
\begin{minipage}{\linewidth}\footnotesize
Each judge scores the same $300$ responses per model in three runs. Reported variances are $0.09315$ for GPT-5.4 mini and $0.07276$ for GPT-6.1 Sol. The last row uses unrounded means and gives percentage-point gains over Direct QA.
\end{minipage}
\end{table}

%% file: tables/reasoning_anlaysis.tex
\begin{table}[t]
\centering
\caption{Clinical rationale evaluation. Reference-step recall is scored on MedCaseReasoning. Physician rows give preference counts from $40$ annotations on $30$ jointly correct cases with $10$ shared cases. Win rates exclude ties.}
\label{tab:reasoning_eval}
\label{tab:human_study_counts}
\small
\setlength{\tabcolsep}{4pt}
\renewcommand{\arraystretch}{1.25}
\begin{tabular}{@{}L{\dimexpr0.32\textwidth-1.6\tabcolsep\relax}C{\dimexpr0.26\textwidth-1.6\tabcolsep\relax}C{\dimexpr0.18\textwidth-1.6\tabcolsep\relax}C{\dimexpr0.10\textwidth-1.6\tabcolsep\relax}C{\dimexpr0.14\textwidth-1.6\tabcolsep\relax}@{}}
\toprule
\multicolumn{1}{@{}l}{\textbf{Metric}}
& \multicolumn{1}{c}{\makecell[c]{\raisebox{-3pt}[0pt][0pt]{\includegraphics[height=12pt,trim=44bp 38bp 27bp 0bp,clip]{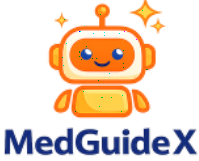}}\hspace{3pt}\textbf{\model (Ours)}}}
& \multicolumn{1}{c}{\makecell[c]{\raisebox{-3pt}[0pt][0pt]{\includegraphics[height=12pt]{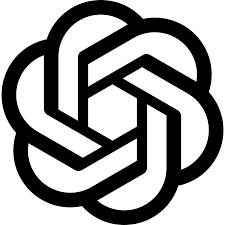}}\hspace{3pt}\textbf{GPT}}}
& \multicolumn{1}{c}{\makecell[c]{\textbf{Tie}}}
& \multicolumn{1}{c}{\makecell[c]{\textbf{Win rate}\\\textbf{\%}}} \\
\midrule
Reference-step recall & \textbf{62.8\%} & 50.8\% & \textcolor{gray}{n/a} & \textcolor{gray}{n/a} \\
\midrule
Evidence faithfulness & 17 & 3 & 20 & 85.0 \\
Reasoning validity & 23 & 6 & 11 & 79.3 \\
Reasoning completeness & 36 & 3 & 1 & 92.3 \\
Reasoning conciseness & 17 & 16 & 7 & 51.5 \\
\midrule
\textbf{Overall} & \textbf{93} & \textbf{28} & \textbf{39} & \textbf{76.9} \\
\bottomrule
\end{tabular}
\end{table}

%% file: appendix.tex
\section*{Contents of Appendix}

This appendix provides additional details supporting the main paper.

\begin{itemize}[leftmargin=*, itemsep=2pt, topsep=2pt]
    \item \textbf{\ref{sec:limitations}. Limitations} summarizes the current limitations of \model and its intended research-use scope.
    \item \textbf{\ref{sec:ethic}. Ethical Statement} discusses data privacy, research use, and clinical safety considerations.
    \item \textbf{\ref{sec:data_details}. Data Preparation Details} describes guideline curation, executable transformation, QA synthesis, and label and source audits.
    \item \textbf{\ref{sec:benchmark_details}. Benchmark Details and Examples} details the evaluation datasets, scoring protocols, licenses, and representative examples.
    \item \textbf{\ref{sec:training_details}. Training Details} reports SFT and RL settings for training \model.
    \item \textbf{\ref{sec:ablation}. Training Strategy Ablation} covers the Direct QA control, training schedules, answer transitions, and held-out guideline evaluation. Appendix~\ref{sec:heldout_prompts} gives its prompts.
    \item \textbf{\ref{sec:human_study}. Human Study Details} describes the physician evaluation design, annotation protocol, scoring, and questionnaire.
    \item \textbf{\ref{sec:guideline_baselines}. Details of Guideline-Based Baselines} explains the implementation of retrieval, in-context, fine-tuning, and RL guideline baselines.
    \item \textbf{\ref{sec:case_study}. Case Study} presents qualitative diagnostic case studies comparing \model with baseline models.
    \item \textbf{\ref{sec:ai_use}. AI Use Statement} describes AI assistance in writing, literature search, and synthetic data generation.
\end{itemize}

\section{Limitations}
\label{sec:limitations}
\model is a research system for clinical reasoning. The benchmark results cover the reported decoding protocol. The held-out guideline test measures action agreement with GPT-6.1 Sol xhigh pseudolabels, and its matched triplets have correlated errors. The physician study describes a small, directionally curated sample. Automated execution checks reproduce the compiled rules, while source fidelity and hidden-state ambiguity require separate assessment. Appendix~\ref{sec:label_audit} reports concrete failures in both areas. Clinical deployment requires broader validation and safety evaluation.

\section{Ethical Statement}
\label{sec:ethic}
This work aims to improve clinical reasoning in medical LLMs by transforming publicly available clinical practice guidelines into structured training supervision. All guideline sources used in this study are publicly accessible and do not contain private patient information. The factual and counterfactual QA instances are synthetically generated from executable guideline functions and are not derived from identifiable patient records.

We acknowledge that medical LLMs may produce incorrect, incomplete, or potentially harmful recommendations, even when trained on guideline-derived data. Therefore, \model is intended for research use only and should not be used as a substitute for professional medical judgment, clinical diagnosis, or treatment decisions. Any deployment in real-world healthcare settings would require rigorous clinical validation, human oversight, bias and safety evaluation, and compliance with applicable medical regulations.

\section{Data Preparation Details}
\label{sec:data_details}

\subsection{Guideline Curation}
Our initial guideline source is the open EPFL MEDITRON clinical guideline collection~\citep{chen2023meditron}, containing $37{,}970$ documents. The released collection is available at \url{https://huggingface.co/datasets/epfl-llm/guidelines}. After curation, we obtain a filtered subset of $841$ CPGs, including $607$ from CDC and $234$ from PubMed. 

\subsection{Executable Transformation}

We use GPT-5.4~\citep{openai2026gpt54} with reasoning effort set to \texttt{xhigh} for recommendation extraction, decision-tree generation, and decision-tree validation.

Starting from $841$ curated documents, we split the corpus into $3{,}127$ recommendation-oriented chunks using a soft limit of $4{,}500$ words per chunk, with at most $4$ chunks per document and $2$ recommendations per chunk. The LLM extractor produces $4{,}750$ recommendation candidates. After validation, we retain $3{,}796$ usable recommendations, discarding $827$ candidates that lack a concrete condition-action structure and removing $127$ near-duplicates with highly similar populations, conditions, and actions.

Each of the $3{,}796$ retained recommendations is converted into a decision tree $T_f$ specifying input variables, decision conditions, and final outputs. After LLM-based validation for completeness, condition clarity, branch coverage, and guideline support, $2{,}800$ trees pass validation.

A deterministic converter translates the $2{,}800$ validated trees into executable Python functions that map input variables $X$ to guideline outputs $f(X)$. We check each function for syntactic correctness, executability on sampled inputs, and consistency with the original decision tree. In total, $2{,}793$ functions pass these checks and form the executable guideline knowledge base. Validation here refers to the automated checks above.

\subsection{QA Synthesis}

At the data level, executing $f$ before and after intervention yields $7{,}993$ interventional scenarios. Of these, $7{,}205$ produce changed outcomes, while $788$ unchanged-output cases are discarded. After applying the same balancing rule used for factual QA, we retain $5{,}028$ counterfactual candidates. Each sample is paired with a reasoning trace generated by GPT-5.4~\citep{openai2026gpt54} with reasoning effort set to \texttt{xhigh} that verbalizes the abduction, intervention, and prediction steps over the executable function $f$. Overall data pipeline statistics are shown in Table~\ref{tab:data_stats}.

\subsection{Label Reproducibility and Source Checks}
\label{sec:label_audit}
\paragraph{Replay and hidden-state identifiability.}
We replay all $4{,}963$ factual labels and the factual and intervened outputs of all $5{,}028$ counterfactual records in the released corpus. Every stored output is reproduced at its stored input. Among $4{,}623$ records with a Boolean hidden variable, exhaustive enumeration finds $1{,}123$ with two hidden values that reproduce the factual output. Both values also produce the same counterfactual output under exact list matching. The sampled hidden state is therefore not uniquely identifiable in these records.

Let $H$ be the legal hidden-variable domain and let
\begin{equation}
\mathcal{H}_{\mathrm{compat}}
=\{h\in H\mid f(X_{\mathrm{obs}},h,x_{\mathrm{int}})=Y_{\mathrm{obs}}\}.
\end{equation}
Single-answer supervision is identifiable when $f(X_{\mathrm{obs}},h,\hat{x}_{\mathrm{int}})$ agrees for every $h\in\mathcal{H}_{\mathrm{compat}}$. A numerical counterexample occurs in the released ICU census rule. Usual capacities of $12$ and $13$ both explain no action at a census of $30$, but raising the census to $36$ triggers the three-times-capacity threshold only for $12$. Numerical probing is not exhaustive, so it provides counterexamples rather than an ambiguity rate.

\paragraph{Action multiplicity.}
In $213$ of $5{,}028$ counterfactual records, the intervention changes only repetitions of identical action strings. Of these $213$ records, $200$ have a Boolean hidden variable. Removing those $200$ records leaves $4{,}423$ Boolean-hidden records. Under comparison of distinct action sets, one remaining record admits different counterfactual actions across compatible hidden values. Exact list reproducibility and action identifiability therefore measure different properties. The reported checkpoints use the original corpus; the audit does not represent a retraining experiment on corrected labels.

\paragraph{Source fidelity and the illustrated rule.}
The diabetes rule in Figure~\ref{fig:pipeline} follows the four diagnostic criteria and repeat-testing condition in the source~\citep{colberg2010exercise}. All three factual and six counterfactual records for this function replay. Repeat testing is attached to a qualifying laboratory criterion and is absent when no criterion triggers. The diagram summarizes the diagnostic condition, while the code preserves the separate actions.

A targeted check of a pediatric urine-collection recommendation~\citep{chishti2010pyelonephritis} finds an exception-precedence error. The function can return both clean-catch and catheterized collection when the source calls for catheterization if reliable clean catch is unavailable. Giving that exception precedence changes a factual or intervened reference in four of the rule's six counterfactual records; two then become action-preserving. Its two factual records are unchanged. This check identifies a specific compilation failure and does not estimate corpus-wide clinical fidelity.

\section{Benchmark Details and Examples}
\label{sec:benchmark_details}
We evaluate \model on four external medical reasoning benchmarks. We use their evaluation sets and do not train on their benchmark training sets.

\begin{itemize}[leftmargin=*, itemsep=0pt, labelsep=5pt, topsep=0pt]
    \item \textbf{MedQA}~\citep{jin2021disease}: A multiple-choice medical exam benchmark based on United States Medical Licensing Examination (USMLE)-style questions. It evaluates whether executable guideline-grounded training transfers to exam-style medical knowledge. We report multiple-choice accuracy on $1{,}273$ test questions.

    \item \textbf{MedCaseReasoning}~\citep{wu2025medcasereasoning}: A long-form open diagnostic reasoning benchmark based on open-access clinical case reports from the PubMed Central Open Access Subset. It evaluates case-based differential diagnosis. We report diagnostic accuracy on the $897$ cases in the test split.

    \item \textbf{MIMIC-CDM-FI}~\citep{hager2024evaluation}: A full-information open clinical decision-making benchmark derived from MIMIC-IV, which is based on electronic health records from Beth Israel Deaconess Medical Center. It evaluates full-information clinical decision making. We report diagnostic accuracy on a fixed class-balanced subset of $1{,}000$ cases.

    \item \textbf{ER-Reason}~\citep{mehandru2025er}: An emergency-department open diagnosis prediction benchmark derived from patient records at a large academic medical center. It evaluates emergency-room diagnosis prediction. We report diagnostic accuracy on a fixed subset of $1{,}000$ cases.
\end{itemize}

Unless otherwise stated, we use greedy decoding with a $1{,}024$-token generation budget and disable Qwen thinking mode. GPT-5.4 mini~\citep{openai2026gpt54mini} with \texttt{xhigh} reasoning effort performs the benchmark diagnosis and reasoning-recall judgments described below. For each benchmark, we describe the task, split, and evaluation protocol, and provide one representative example.

\subsection{MedQA}
\label{sec:medqa_protocol}
MedQA is a USMLE-style multiple-choice medical QA benchmark. We evaluate on the US test split of $1{,}273$ questions. Given a question stem and five options A through E, the model selects the single best answer after a chain-of-thought reasoning step. Answer extraction is fully deterministic: a regex-based parser strips reasoning blocks and extracts the final option letter or option text, with no LLM judge involved. We report accuracy as the fraction of correctly answered questions.

\begin{examplebox}{MedQA: USMLE-style multiple-choice QA}
\footnotesize
\exrole{System}
You are a medical expert. Answer the multiple-choice question, thinking step by step before choosing one option from A to E.

\exrole{User}
\textit{Context.} A 65-year-old man with fever, productive cough, pleuritic chest pain, and right lower-lobe consolidation on chest x-ray. Gram stain shows lancet-shaped gram-positive diplococci.\\
\textit{Question.} Which option is the single best answer?\\
\textit{Options.} A: \textit{Mycoplasma pneumoniae} \quad B: \textit{Streptococcus pneumoniae} \quad C: \textit{Pseudomonas aeruginosa} \quad D: \textit{Pneumocystis jirovecii} \quad E: \textit{Histoplasma capsulatum}

\exout{Reference diagnosis}
B

\exout{Evaluation}
The deterministic parser extracts option B from the model output and matches it against the gold answer B. The prediction is counted as correct.
\end{examplebox}

\subsection{MedCaseReasoning}
\label{sec:mcr_protocol}
MedCaseReasoning evaluates long-form diagnostic reasoning from open-access clinical case reports. Each example provides a clinical case prompt, a clinician-authored reasoning trace, and a final diagnosis. We evaluate on the official test split of $897$ cases and use the official prompt template, which requests reasoning inside \texttt{<think>} tags and the diagnosis inside \texttt{<answer>} tags. We report $1$-pass diagnostic accuracy under the official LLM-as-judge protocol, in which the judge compares the predicted and gold diagnoses and counts synonyms, abbreviations, and close medical paraphrases as correct. We additionally report reasoning recall, computed by a separate judge that checks, for each clinician-authored reasoning step, whether the model's reasoning explicitly or implicitly covers it.

\begin{examplebox}{MedCaseReasoning: long-form case diagnosis}
\footnotesize
\exrole{User}
Please reason through the following clinical case and predict the final diagnosis, returning reasoning inside \texttt{<think>} tags and the diagnosis inside \texttt{<answer>} tags.\\[2pt]
\textit{Clinical case.} A middle-aged patient presents with severe episodic headaches, palpitations, sweating, and paroxysmal hypertension. Abdominal imaging reveals an adrenal mass. Plasma metanephrines are markedly elevated.

\exout{Reference diagnosis}
Pheochromocytoma

\exout{Evaluation}
The judge compares the predicted diagnosis against the gold diagnosis pheochromocytoma and finds them equivalent. Reasoning recall is computed over the clinician-authored reasoning steps; in this case all reference steps are covered, giving a recall of $1.0$.
\end{examplebox}

\subsection{MIMIC-CDM-FI}
\label{sec:mimic_protocol}
MIMIC-CDM-FI is a full-information clinical decision-making benchmark derived from the MIMIC-IV-Ext CDM dataset. We evaluate on a fixed $1{,}000$-case test split balanced across four acute abdominal pathologies, with $250$ cases each of appendicitis, cholecystitis, pancreatitis, and diverticulitis. The model receives all relevant patient information upfront, including history, physical examination, selected laboratory results, and abdominal imaging reports, and is asked to output a single final diagnosis of the most severe pathology with no further explanation. Predictions are scored by an LLM-as-judge that identifies the primary acute abdominal diagnosis. Synonyms, spelling variants, abbreviations, and clinically equivalent paraphrases count as correct; comorbidities or full billing-code matches are not required.

\begin{examplebox}{MIMIC-CDM-FI: full-information clinical decision making}
\footnotesize
\exrole{System}
You are a medical AI assistant. Based on the provided information, give a single final diagnosis of the most severe pathology, with no further information.

\exrole{User}
Provide the most likely final diagnosis of the following patient.\\[2pt]
\textit{History.} Acute right lower quadrant abdominal pain, nausea, anorexia, low-grade fever.\\
\textit{Examination.} Tenderness at McBurney's point with guarding.\\
\textit{Laboratory.} Elevated white blood cell count and C-reactive protein.\\
\textit{Imaging.} CT abdomen shows a dilated appendix with wall thickening and periappendiceal fat stranding.

\exout{Reference diagnosis}
Acute appendicitis

\exout{Evaluation}
The judge identifies the model's response as a match for the gold diagnosis acute appendicitis. The prediction is counted as correct.
\end{examplebox}

\subsection{ER-Reason}
\label{sec:er_protocol}
ER-Reason evaluates clinical reasoning in emergency-department settings; we use Task~4, final ED diagnosis prediction. We evaluate on a deterministic $1{,}000$-record split. Each record provides age, sex, chief complaint, the current ED presentation, and a set of clinical notes spanning discharge summary, progress notes, history and physical, imaging, and consult. The model receives all notes as context and outputs a single ED diagnosis in free text. Our primary metric is LLM-judge accuracy, with the judge counting synonyms, abbreviations, wording differences, and clinically equivalent specificity as correct. The official Task~4 evaluator also provides a CMS ICD-10-CM crosswalk that yields exact match, normalized match, token F1, clinical-cluster, body-system, ICD, and HCC accuracy. We do not adopt the crosswalk-based HCC accuracy as our primary metric because the crosswalk maps only about $30\%$ of the gold and predicted diagnoses to an HCC category. The resulting score is computed over a small and non-representative subset of cases, whereas LLM-judge accuracy scores every case and better reflects clinically correct predictions.

\begin{examplebox}{ER-Reason (Task 4): emergency-department diagnosis}
\footnotesize
\exrole{System}
You are an experienced ED physician. Predict the most likely diagnosis for the current ED visit, using all available clinical notes as full information.

\exrole{User}
\textit{ER-Reason Task 4: Final Diagnosis.} Age 72, Female. Chief complaint: shortness of breath.\\
\textit{Clinical notes.} History of heart failure; worsening dyspnea, orthopnea, bilateral leg edema, elevated BNP, pulmonary vascular congestion on chest x-ray.\\
\textit{Current ED presentation.} Tachypneic with hypoxia and bibasilar crackles; no fever or focal infiltrate.\\
Output a single CMS-standardized diagnosis as text, with no explanation.

\exout{Reference diagnosis}
Acute congestive heart failure

\exout{Evaluation}
The judge compares the model's prediction acute decompensated heart failure against the gold acute congestive heart failure and counts them as clinically equivalent. The prediction is counted as correct.
\end{examplebox}

\begin{table}[t]
    \caption{Summary of the four evaluation benchmarks. \#~Eval is the number of evaluated examples. MedQA uses a fully deterministic answer parser with no LLM judge; the other three use an LLM-as-judge (\texttt{gpt-5.4-mini}, reasoning effort \texttt{xhigh}, temperature $0$) for semantic diagnosis matching. ER-Reason also has an official CMS ICD-10-CM crosswalk evaluator available, but it is not used as the primary metric (see Appendix~\ref{sec:er_protocol}).}
    \label{tab:benchmark_summary}
    \centering
    \small
    \setlength{\tabcolsep}{4pt}
    \renewcommand{\arraystretch}{1.2}
    \begin{tabular*}{\columnwidth}{@{\extracolsep{\fill}} l c c c @{}}
        \toprule
        \multicolumn{1}{c}{\textbf{Benchmark}} & \textbf{Split} & \textbf{\# Eval} & \textbf{Scoring} \\
        \midrule
        MedQA            & US test    & $1{,}273$ & parser \\
        MedCaseReasoning & test       & $897$     & LLM judge \\
        MIMIC-CDM-FI     & test\_1000 & $1{,}000$ & LLM judge \\
        ER-Reason        & test\_1000 & $1{,}000$ & LLM judge \\
        \bottomrule
    \end{tabular*}
\end{table}

\paragraph{Licenses and access.}
We confirm that all four evaluation benchmarks are used in accordance with their original licenses, and only for non-commercial academic research. MedQA is released under the MIT license via the official GitHub repository.\footnote{\url{https://github.com/jind11/MedQA}} MedCaseReasoning's code is released under the MIT license and its dataset is released under CC-BY 4.0, derived from the PubMed Central Open Access Subset.\footnote{\url{https://github.com/kevinwu23/Stanford-MedCaseReasoning}}\textsuperscript{,}\footnote{\url{https://huggingface.co/datasets/zou-lab/MedCaseReasoning}} MIMIC-CDM-FI is distributed under the PhysioNet Credentialed Health Data License 1.5.0 with an accompanying Data Use Agreement, and access requires completion of the CITI Data or Specimens Only Research training; we accessed it as credentialed PhysioNet users and use it only for evaluation in this study.\footnote{\url{https://physionet.org/content/mimic-iv-ext-cdm/}} ER-Reason is distributed under the more restrictive PhysioNet Contributor Review Health Data License 1.5.0, which additionally requires per-study review by the dataset contributors; we accessed it under this credentialed and reviewed access process and use it only for evaluation.\footnote{\url{https://physionet.org/content/er-reason/1.0.0/}} We do not redistribute any patient-level records from MIMIC-CDM-FI or ER-Reason, and all evaluation outputs reported in this paper are aggregate metrics rather than raw clinical content.

\paragraph{Summary.}
Table~\ref{tab:benchmark_summary} summarizes the four benchmarks. MedQA isolates broad exam-style medical knowledge with deterministic scoring; MedCaseReasoning targets long-form differential diagnosis with both accuracy and reasoning-recall judging; MIMIC-CDM-FI measures full-information acute abdominal decision making on a class-balanced split; and ER-Reason measures ED diagnosis prediction with semantic diagnosis judging.

\section{Training Details}
\label{sec:training_details}

\paragraph{Training setup.}
We train \model-4B and \model-9B from Qwen3.5-4B and Qwen3.5-9B, respectively. Both models use the same post-training recipe: supervised fine-tuning on the guideline-derived factual and counterfactual QA mixture, followed by GRPO on factual guideline QA prompts. Unless otherwise stated, the details below describe the 9B run; the 4B run follows the same data construction and optimization setup with the corresponding Qwen3.5-4B backbone.

\paragraph{Supervised fine-tuning.}
The SFT stage is designed to teach the model how to express the executable guideline logic in natural-language reasoning. We train on the balanced mixture of factual QA and counterfactual QA examples described in Section~\ref{sec:data}. For factual examples, the target response contains the guideline-prescribed output and a rationale verbalizing the executed path $\pi_f(X)$. For counterfactual examples, the target response follows the abduction--intervention--prediction structure, requiring the model to infer hidden variables, apply the intervention, and predict the new guideline output. This stage therefore serves as a cold start for both the answer format and the reasoning pattern used in later RL.

All SFT runs use LoRA adaptation on all linear layers, with rank $16$ and $\alpha=32$, bfloat16 precision, and gradient checkpointing. We train with verl~\citep{sheng2025hybridflow} for $5$ epochs, using a learning rate of $1\times10^{-5}$, a cosine learning-rate schedule with a warmup ratio of $0.03$, maximum sequence length $2048$, and global batch size $64$. For the 9B model, the SFT stage runs for approximately $16$ hours on $8\times$ RTX 5090 GPUs, and the resulting checkpoint is used to initialize the RL stage.

\paragraph{Reinforcement learning.}
After SFT, we apply GRPO~\citep{shao2024deepseekmath} using the same verl training stack. The final \model configuration uses factual RL, selected by the ablation study in Section~\ref{sec:training_ablations}. The RL prompts reuse the $4{,}963$ factual prompts from the guideline-derived SFT corpus. The benchmark training sets are excluded from post-training.

During RL, each prompt is sampled with multiple rollouts and scored by the reward function described in Section~\ref{sec:training}. The reward first enforces the required reasoning-before-answer format, then checks whether the parsed final guideline output matches the oracle output produced by the executable function. The original counterfactual verifier is described below. The paired Direct QA comparison uses the common answer-scoring interface described in Appendix~\ref{sec:direct_qa_protocol}.

The RL stage is initialized from the SFT checkpoint and trained for one epoch with learning rate $5\times10^{-6}$. We use a maximum prompt length of $1536$, maximum response length of $1024$, $24$ rollouts per prompt, sampling temperature $1.0$, top-$p=1.0$, one PPO update epoch, entropy coefficient $0.0$, and KL coefficient $0.005$ with the low-variance KL estimator. We evaluate and save checkpoints every $10$ training steps, selecting the final checkpoint on QA from held-out guidelines. The validation split is defined at the source-guideline level, with validation guidelines excluded from SFT and RL training. This checkpoint-selection step is distinct from the held-out guideline evaluation in Section~\ref{sec:decision_behavior}. For the 9B model, the RL stage runs for approximately $33$ hours on $8\times$ RTX 5090 GPUs, and the resulting checkpoint is the final \model-9B checkpoint.

\subsection{Counterfactual Reward Scope}
\label{sec:cf_reward}
The released counterfactual verifier parses an inferred hidden state $\hat{X}_{\mathrm{hid}}$ and action $\hat{Y}_{\mathrm{cf}}$. Its answer reward is
\begin{equation}
\label{eq:counterfactual_correctness}
\begin{aligned}
r_{\mathrm{CF}}(o)={}&
\mathds{1}[\hat{X}_{\mathrm{hid}}=X_{\mathrm{hid}}^{\star}]
\cdot\mathds{1}[\hat{Y}_{\mathrm{cf}}=Y_{\mathrm{cf}}^{\star}]\\
&\cdot\mathds{1}[f(X_{\mathrm{obs}},\hat{X}_{\mathrm{hid}},x_{\mathrm{int}})=Y_{\mathrm{obs}}]\\
&\cdot\mathds{1}[f(X_{\mathrm{obs}},\hat{X}_{\mathrm{hid}},\hat{x}_{\mathrm{int}})=Y_{\mathrm{cf}}^{\star}].
\end{aligned}
\end{equation}
Here the starred state and action are the stored sampled state and execution target. The first factor rejects an alternative compatible hidden value in all $1{,}123$ nonunique Boolean cases described in Appendix~\ref{sec:label_audit}. The released SFT builder also uses the sampled hidden value in counterfactual targets. These checks enforce sampled-state recovery. Accepting any compatible completion is appropriate when all such completions agree on the intervened action; competing actions require ambiguity-aware targets or exclusion from single-answer supervision. Final \model GRPO uses factual prompts and Equation~\ref{eq:factual_correctness}.

\section{Training Strategy Ablation}
\label{sec:ablation}
\subsection{Matched Direct QA Control}
\label{sec:direct_qa_protocol}
Both supervision arms share source guidelines, patient questions, student inputs, training volume, and optimization budgets. Both use GPT-5.4 with reasoning effort set to \texttt{xhigh} for text generation. The released student format attaches the guideline function to the question. The paired control fixes the full student input across arms. The direct generator reads the guideline and question and produces an answer and rationale. The executable arm determines the target action by running the function and generates a rationale from that execution. Each arm is trained against its own reference answers. Both use the same format and final-answer scoring interface, including the CF configurations in this paired comparison. Equation~\ref{eq:counterfactual_correctness} documents the released sampled-state verifier and is separate from this common scoring rule. The shared cases were constructed by the guideline pipeline, so this control evaluates supervision generation on those cases.

\subsection{Training Schedules and Stage Effects}
\label{sec:ablation_schedules}
Figure~\ref{fig:ablation}a compares $15$ data and training schedules at 9B. F denotes factual QA, CF counterfactual QA, and Mix an equal mixture. SFT-only runs apply one of five schedules, factual, counterfactual, factual then counterfactual, counterfactual then factual, or mixed. RL-only runs start from the backbone and use one of the four unmixed schedules. The six SFT-plus-RL runs initialize RL from the corresponding SFT checkpoint. Arrows specify successive data phases within SFT or RL.

Mixed SFT gives gains of $1.6$ and $2.3$ points for Direct QA and $2.1$ and $3.1$ for executable supervision on MedCaseReasoning and MedQA. Factual RL alone gives $4.7$ and $3.0$ points for Direct QA and $6.2$ and $3.7$ for executable supervision. Mixed SFT followed by factual RL performs best in both arms, reaching $6.5$ and $5.6$ points for Direct QA and $8.7$ and $6.9$ for executable supervision. Sequential SFT schedules perform worse than mixed SFT on these two benchmarks. Factual RL also exceeds CF RL within both arms. The full numerical grid is displayed in Figure~\ref{fig:ablation}a.

\subsection{Answer Transition Counts}
\label{sec:transition_details}
For each benchmark item, Corrected means that the backbone is wrong and \model is correct; Regressed means the reverse; Unchanged includes both jointly correct and jointly incorrect answers. Table~\ref{tab:transition} gives counts underlying Figure~\ref{fig:ablation}c. The denominators are $897$, $1{,}273$, $1{,}000$, and $1{,}000$ for the four benchmarks. Corrected minus Regressed is the net gain in correct answers.
\input{tables/paired_analysis}

\subsection{Held-Out Guideline Decision Evaluation}
\label{sec:heldout_eval_protocol}
The source guidelines are drawn from the EPFL MEDITRON clinical guideline collection~\citep{chen2023meditron}, which also supplies the training corpus. Evaluation is held out at the source-document level. These guideline documents and their derived cases are excluded from SFT and RL.

Each of the $100$ triplets contains an original case and two single-variable edits. One edit preserves the guideline action and the other changes it. All clinically relevant case variables are supplied with the guideline text, and each case version is queried independently. This evaluation therefore requires applying the supplied rule to complete inputs. GPT-6.1 Sol~\citep{openai2026gpt61sol} with reasoning effort set to \texttt{xhigh} generates the case text and action pseudolabels. These labels are used directly for scoring. The three models receive identical guideline passages and case prompts.

Table~\ref{tab:guideline_decisions} reports $176$, $193$, and $205$ correct actions out of $300$ for Qwen3.5-9B, Direct QA 9B, and \model-9B. The same $100$ originals enter both paired comparisons. A preserving or changing pair earns one point only when both responses match their respective action targets.

For written paths, each judge receives the guideline, case and question, guideline-based reference path, and one anonymous response. Credit requires identifying the decisive facts, applying the relevant rule and exceptions, and deriving the correct action. A correct action without supporting reasoning earns zero. Equivalent valid paths receive credit; an ambiguous source or conflicting reference is flagged for review. Each judge's reported rate averages three binary scoring runs on the same $300$ responses per model. Appendix~\ref{sec:heldout_prompts} gives the prompt specifications and full judge template.

Medical expert review assessed whether GPT-6.1 Sol had correctly assigned a pass or fail verdict to each of $50$ sampled written paths. The expert assessment agreed with $47$ verdicts and disagreed with $3$, yielding $94\%$ agreement.

\section{Human Study Details}
\label{sec:human_study}
\begin{figure}[!t]
\centering
\small
\begin{tcolorbox}[colback=gray!4,colframe=gray!55,boxrule=0.4pt,arc=2pt,
left=6pt,right=6pt,top=5pt,bottom=5pt,enhanced]
\textbf{Case 027\quad Clinical question}\\[2pt]
A 29-year-old Indigenous Australian man presented with severe epigastric pain radiating to his back. His history included type~2 diabetes, obesity, hypertriglyceridemia, and prior hypertriglyceridemia-induced pancreatitis. He denied alcohol use and appeared clinically dehydrated. Post-resuscitation triglycerides were 24.1~mmol/L, lipase 119~U/L, and glucose 21.3~mmol/L, with no ketonuria. Contrast-enhanced abdominal CT at 24 hours showed fatty liver and no features of pancreatitis.

\smallskip
\textbf{Reference diagnosis}\quad Hyperlipemic abdominal crisis
\end{tcolorbox}

\begin{minipage}[t]{0.49\textwidth}
\begin{tcolorbox}[colback=gray!2,colframe=gray!55,boxrule=0.4pt,arc=2pt,
left=6pt,right=6pt,top=5pt,bottom=5pt,title={Response A},coltitle=black,
fonttitle=\bfseries\small,equal height group=physicianresponses,enhanced]
\textbf{Reasoning}\\[2pt]
Severe epigastric pain with very high triglycerides suggests a lipid-related abdominal syndrome, but lipase was only mildly elevated and contrast CT at 24 hours showed no pancreatitis, failing to meet diagnostic criteria. The sample was lipemic, yet even accounting for assay interference, the absence of imaging changes supports a non-pancreatitic hyperlipemic abdominal crisis. [\ldots]

\smallskip
\textbf{Final diagnosis}\\[2pt]
Hyperlipemic abdominal crisis [\ldots]
\end{tcolorbox}
\end{minipage}\hfill
\begin{minipage}[t]{0.49\textwidth}
\begin{tcolorbox}[colback=gray!2,colframe=gray!55,boxrule=0.4pt,arc=2pt,
left=6pt,right=6pt,top=5pt,bottom=5pt,title={Response B},coltitle=black,
fonttitle=\bfseries\small,equal height group=physicianresponses,enhanced]
\textbf{Reasoning}\\[2pt]
[\ldots] However, his lipase is only mildly elevated (119 U/L), and contrast-enhanced CT performed 24 hours after symptom onset shows no features of pancreatitis, only fatty liver. [\ldots] Given the absence of imaging evidence of pancreatitis and the presence of severe abdominal pain in the setting of extreme hypertriglyceridemia, the most plausible explanation is a hyperlipemic abdominal crisis, which can cause severe abdominal pain without overt pancreatitis on imaging. [\ldots]

\smallskip
\textbf{Final diagnosis}\\[2pt]
Hyperlipemic abdominal crisis
\end{tcolorbox}
\end{minipage}

\begin{tcolorbox}[colback=white,colframe=gray!55,boxrule=0.4pt,arc=2pt,
left=6pt,right=6pt,top=5pt,bottom=5pt,enhanced]
\textbf{Case 027\quad Pairwise judgments}\\[2pt]
\footnotesize Select Response~A, Response~B, or Same for each dimension.
\par\smallskip
\setlength{\tabcolsep}{3pt}
\renewcommand{\arraystretch}{1.35}
\begin{tabular}{@{}p{0.28\linewidth}p{0.45\linewidth}p{\dimexpr0.27\linewidth-12pt\relax}@{}}
\textbf{Dimension} & \textbf{Judgment} & \textbf{Comment} \\
\hline
Evidence Faithfulness &
$\square$~Response~A \hfill $\square$~Response~B \hfill $\square$~Same &
\fbox{\parbox[c][1.2em][c]{\dimexpr\linewidth-2\fboxsep-2\fboxrule\relax}{\centering\scriptsize\textcolor{gray}{Optional elaboration}}} \\
Reasoning Validity &
$\square$~Response~A \hfill $\square$~Response~B \hfill $\square$~Same &
\fbox{\parbox[c][1.2em][c]{\dimexpr\linewidth-2\fboxsep-2\fboxrule\relax}{\centering\scriptsize\textcolor{gray}{Optional elaboration}}} \\
Reasoning Completeness &
$\square$~Response~A \hfill $\square$~Response~B \hfill $\square$~Same &
\fbox{\parbox[c][1.2em][c]{\dimexpr\linewidth-2\fboxsep-2\fboxrule\relax}{\centering\scriptsize\textcolor{gray}{Optional elaboration}}} \\
Reasoning Conciseness &
$\square$~Response~A \hfill $\square$~Response~B \hfill $\square$~Same &
\fbox{\parbox[c][1.2em][c]{\dimexpr\linewidth-2\fboxsep-2\fboxrule\relax}{\centering\scriptsize\textcolor{gray}{Optional elaboration}}}
\end{tabular}
\end{tcolorbox}
\caption{Layout of the blinded physician questionnaire with an archived case and response pair. The form includes the clinical question, reference diagnosis, two anonymized responses, and four judgments with optional comments. One response is from \model-9B and one from GPT-5.0, with presentation order randomized per case. Both responses give the reference diagnosis. The case is summarized and response excerpts are verbatim, with omissions marked by ellipses.}
\label{fig:human_study_example}
\end{figure}

\paragraph{Study design.}
We conduct a blinded pairwise comparison study to assess the clinical reasoning quality of \model-9B against GPT-5.0. The study covers $30$ clinical cases curated from the subset of MedCaseReasoning on which both models produce a correct final diagnosis. The selection record explicitly favors \model rationales under the four rating dimensions. It also filters for reference-reasoning recall and final-answer overlap at least as high as the comparator, and for \model rationale length of at least $500$ characters. The resulting preferences describe these selected cases. For each case, an annotator is shown the case prompt, the ground-truth diagnosis, and two anonymized responses, and judges which response is better along four reasoning dimensions. Model identities are hidden, and the left/right presentation order of the two responses is independently randomized per case so that the annotator cannot infer which system produced which response.

\paragraph{Annotators and case allocation.}
Annotations are provided by two practicing physicians. Each physician annotates $20$ cases, with $10$ cases shared between them. Concretely, physician 1 annotates cases $1$--$20$ and physician 2 annotates cases $11$--$30$, so that cases $11$--$20$ are double annotated. This design covers all $30$ cases with $40$ total annotations while keeping per-annotator load manageable, with $10$ cases receiving two ratings. Annotators are not told the identity, size, or training procedure of either model.

\paragraph{Evaluation dimensions.}
The four questionnaire dimensions are evidence faithfulness, reasoning validity, reasoning completeness, and reasoning conciseness. The evaluation criteria are as follows.
\begin{itemize}[leftmargin=*, itemsep=2pt, topsep=2pt]
    \item \textbf{Evidence Faithfulness.} Which response is more faithful to the evidence provided in the case? Prefer the response that grounds its reasoning in the stated history, symptoms, labs, imaging, and pathology, and penalize unsupported claims, invented findings, or conclusions that go beyond the provided evidence.
    \item \textbf{Reasoning Validity.} Which response has more valid clinical reasoning? Prefer the response whose diagnostic logic is medically sound, internally consistent, and appropriate for the task, and penalize incorrect causal links, medically implausible interpretations, or contradictions.
    \item \textbf{Reasoning Completeness.} Which response provides a more complete reasoning process? Prefer the response that covers the key positive and negative evidence, necessary intermediate steps, and important differential considerations, and penalize responses that skip critical evidence or jump to a diagnosis without adequate justification.
    \item \textbf{Reasoning Conciseness.} Which response is more concise and focused while still being clinically useful? Prefer responses that are clear, organized, and free of unnecessary repetition or irrelevant detail, without rewarding brevity that omits important reasoning.
\end{itemize}

\paragraph{Annotation protocol.}
The study uses a structured questionnaire with a clinical question, two blinded responses, and one judgment per evaluation dimension. The reported outcomes are preference for \model, preference for GPT-5.0, or Same. Free-text comments are optional.

\paragraph{Scoring.}
After collection, the blinded A/B labels are decoded back to model identities using a held-out mapping that is never exposed during annotation. For each dimension we report the \emph{non-tie win rate} of \model-9B over GPT-5.0, defined as the number of annotations in which \model-9B is preferred divided by the total number of non-tie annotations,
\begin{equation*}
\text{Win Rate}
=
\frac{n_{\text{\model-9B}}}{n_{\text{\model-9B}} + n_{\text{GPT-5.0}}},
\end{equation*}
where $n_{\text{\model-9B}}$ and $n_{\text{GPT-5.0}}$ are the counts of \model-9B-preferred and GPT-5.0-preferred judgments, respectively. Same judgments are reported and excluded from the non-tie denominator. The aggregate \emph{Overall} win rate in Table~\ref{tab:reasoning_eval} is computed by pooling all non-tie judgments across the four dimensions and all $40$ annotations.

\paragraph{Per-dimension judgment counts.}
Table~\ref{tab:human_study_counts} reports reasoning recall and the raw physician judgment counts underlying the non-tie win rates. \model-9B is preferred on every dimension, with the strongest margins on reasoning completeness ($36$ vs $3$) and evidence faithfulness ($17$ vs $3$). Reasoning conciseness is the most contested dimension, with $17$ preferences for \model-9B, $16$ for GPT-5.0, and $7$ ties.

\paragraph{Questionnaire example.}
Figure~\ref{fig:human_study_example} shows the questionnaire layout with an archived case and response pair, including all four rating dimensions, response choices, and comment fields.

\section{Details of Guideline-Based Baselines}
\label{sec:guideline_baselines}

Table~\ref{tab:guideline_baselines} reports the full accuracy results underlying Figure~\ref{fig:ablation}b.
\input{tables/guideline_baselines}

\subsection{Retrieval-Augmented Guideline Prompting}
\label{sec:baseline_rag}
We evaluate an inference-time retrieval baseline that uses the same raw guideline chunks as external knowledge. We build a TF-IDF retriever over the $3{,}127$ raw CPG chunks, using each chunk's topic, section path, and guideline text as the retrieval corpus. For each benchmark instance, we form a query from the test input only: the question stem and answer choices for MedQA, the case prompt for MedCaseReasoning, the patient information and clinical fields for MIMIC-CDM-FI, and the emergency-department presentation and available notes for ER-Reason. Gold labels are never used in retrieval. We retrieve the top-$3$ chunks, truncate each retrieved snippet to $700$ characters, and prepend them to the original benchmark prompt with an instruction that the snippets may or may not be relevant. No model parameters are updated. This baseline obtains $31.88\%$, $74.71\%$, $82.50\%$, and $27.40\%$ on the four benchmarks, with an average accuracy of $54.12\%$.

\subsection{In-Context Guideline Demonstrations}
\label{sec:baseline_icl}
We also evaluate an in-context learning baseline that provides guideline-derived examples at inference time without parameter updates. We select three factual QA examples from the guideline-derived factual QA set. Each demonstration contains a patient/guideline query, a short reasoning trace, and the corresponding guideline output. These three demonstrations are fixed across all test instances and are prepended before the original benchmark prompt, with an instruction that they are demonstrations only and are not facts about the target case. Unlike RAG, this baseline does not retrieve case-specific guideline passages; it tests whether a small number of guideline-style demonstrations can induce the desired reasoning behavior in the base model. This baseline obtains $33.11\%$, $74.94\%$, $80.10\%$, and $28.20\%$ on the four benchmarks, with an average accuracy of $54.09\%$.

\subsection{CPGPrompt}
\label{sec:baseline_cpgprompt}
We adapt CPGPrompt~\citep{deng2026cpgprompt} as a train-free decision-tree prompting baseline. For each benchmark instance, we first retrieve the top-$3$ guideline recommendations from the validated recommendation corpus using TF-IDF retrieval. We then prompt Qwen3.5-9B to construct a compact CPGPrompt-style decision tree as a Python dictionary literal, including the selected recommendation ID, yes/no decision nodes, a final rationale, and a final answer. The generated tree is parsed, normalized, and executed locally by a deterministic Python traverser; the reached terminal action or top-level final answer is then scored under the same benchmark-specific evaluation protocol as the main experiments. This baseline does not use our precompiled executable functions and does not update model parameters. It obtains $14.16\%$, $59.15\%$, $67.60\%$, and $19.90\%$ on the four benchmarks, with an average accuracy of $40.20\%$.

\subsection{Fine-tuning with CPG}
\label{sec:baseline_raw_cpg}
We include a text-only fine-tuning baseline that trains Qwen3.5-9B directly on raw CPG text. The training data are constructed from guideline chunks, where the model is asked to continue or reproduce the raw guideline content under a medical guideline system prompt. This baseline uses only unstructured guideline text and does not use extracted recommendations, decision trees, executable Python functions, generated factual/counterfactual QA pairs, or verifier-based rewards. Its goal is to test whether exposure to guideline text alone is sufficient to improve clinical reasoning. It obtains $28.87\%$, $65.04\%$, $78.40\%$, and $21.50\%$ on the four benchmarks, respectively. The results indicate that raw guideline text alone can lead to negative transfer compared with the Qwen3.5-9B backbone.

\subsection{RL with CPG-Derived Process Rewards}
\label{sec:baseline_process_reward}
Finally, we evaluate an RL baseline that adds an LLM guideline-consistency score to the factual answer reward. This baseline starts from the same mixed factual/counterfactual SFT checkpoint and applies GRPO on factual guideline QA prompts. The reward function reuses the factual CoT format reward and final-answer reward, then adds an LLM-as-a-judge process reward: for each sampled response, we retrieve the most relevant validated guideline recommendation, using an exact CPG identifier when available and otherwise lexical overlap retrieval, and ask the judge whether the model's reasoning process is consistent with the retrieved guideline. The judge is GPT-5.4 mini with reasoning effort set to \texttt{xhigh}. The final reward is the sum of the format reward, answer reward, and process-consistency reward. The comparison isolates the additional process judge from the shared mixed SFT initialization and factual answer reward. It obtains $37.68\%$, $75.73\%$, $83.40\%$, and $27.80\%$ on the four benchmarks, with an average accuracy of $56.15\%$.
\section{Case Study}
\label{sec:case_study}

We provide a controlled qualitative comparison on MedCaseReasoning, a long-form diagnostic reasoning benchmark in which each model receives the same clinical case and must produce a final diagnosis. To make the comparison directly auditable, we select three cases where \model, as shown in this section, is judged correct while both GPT-5.0 and MedReason-8B are judged incorrect. For each case, we preserve the complete clinical case, the complete raw model response, and the verbatim judge rationale for each prediction.

Green boxes indicate correct \model outputs, and red boxes indicate incorrect baseline outputs. The examples illustrate diagnostic distinctions in the selected outputs, with aggregate performance reported in Table~\ref{tab:main_results}.

\subsection{Case 1: Central Serous Chorioretinopathy}

\begin{tcolorbox}[breakable, enhanced, colback=blue!3, colframe=blue!55, boxrule=0.35pt, arc=1.5pt, left=3pt, right=3pt, top=3pt, bottom=3pt, title={\textbf{Task Prompt}}, fonttitle=\small]
{\scriptsize\raggedright
Please reason through the following clinical case and predict the final diagnosis. Do not use tables or bullet points. Return your answer in the following format: \texttt{<think>} [Reasoning here] \texttt{</think>} \texttt{<answer>} [Final diagnosis here] \texttt{</answer>}.
}
\end{tcolorbox}

\begin{tcolorbox}[breakable, enhanced, colback=gray!3, colframe=gray!55, boxrule=0.35pt, arc=1.5pt, left=3pt, right=3pt, top=3pt, bottom=3pt, title={\textbf{Full Clinical Case}}, fonttitle=\small]
{\scriptsize\raggedright
A 42-year-old previously healthy woman presented with a 4-day history of blurring of vision and visual loss in her left eye, associated with mild ocular pain. She denied systemic symptoms, drug use, and had no significant family, medical, or ocular history, except for a type-A behavioral pattern. She had been diagnosed elsewhere with presumed unilateral Vogt--Koyanagi--Harada disease and referred for further evaluation prior to treatment.\par
\smallskip\par
On examination, best-corrected visual acuity was 20/20 in the right eye and 20/40 in the left. Cycloplegic refraction was +0.50 diopters in the right eye and -1.00 diopters in the left. Intraocular pressures were normal bilaterally. Slit-lamp biomicroscopy showed a clinically shallow anterior chamber in the left eye and +1 cells in the anterior vitreous; there was no anterior chamber inflammation. Funduscopic examination of the left eye revealed large areas of neurosensory retinal elevation consistent with subretinal fluid involving the peripapillary region and extending beyond the posterior pole; the right eye fundus was unremarkable.\par
\smallskip\par
Spectral-domain optical coherence tomography of the left eye confirmed serous detachment of the neurosensory retina in the macular, peripapillary, and juxtapapillary regions. Optical biometry demonstrated that the anterior chamber depth was 0.62 mm shallower in the left eye than in the right, and axial length was 0.16 mm shorter in the left eye.
}
\end{tcolorbox}

\noindent\textbf{Model Outputs.}

\begin{tcolorbox}[breakable, enhanced, colback=green!3, colframe=green!55, boxrule=0.35pt, arc=1.5pt, left=3pt, right=3pt, top=3pt, bottom=3pt, title={\textbf{MedGuideX (correct)}}, fonttitle=\small]
{\scriptsize\raggedright
\textbf{Prediction:} Idiopathic central serous chorioretinopathy\par
\textbf{Judge reason:} Idiopathic central serous chorioretinopathy is the same diagnosis as central serous chorioretinopathy; the added idiopathic qualifier does not change the condition.\par
\medskip\noindent\textbf{Raw response:}\par
The patient is a 42-year-old woman with acute unilateral visual loss and blurring in the left eye, mild ocular pain, and a shallow anterior chamber. She was previously diagnosed with Vogt-Koyanagi-Harada disease, but the current presentation is atypical for VKH, which typically presents with bilateral uveitis, meningismus, and systemic symptoms such as hearing loss or skin changes. Here, the anterior chamber inflammation is minimal (only +1 cells), and there is no significant vitritis or posterior uveitis on exam. Instead, the key findings are a shallow anterior chamber, reduced axial length, and extensive serous retinal detachment involving the macula, peripapillary region, and beyond the posterior pole. The shallow anterior chamber and shorter axial length in the affected eye suggest a structural abnormality rather than an inflammatory process. These features are characteristic of a primary vitreoretinal disorder, specifically idiopathic central serous chorioretinopathy (CSC), which can present with serous retinal detachment and sometimes a shallow anterior chamber due to ciliary body effusion or altered aqueous dynamics. However, the combination of shallow anterior chamber, reduced axial length, and extensive serous detachment is more consistent with a diagnosis of idiopathic central serous chorioretinopathy with associated anterior chamber shallowing, possibly due to ciliary body effusion or a variant of CSC. Given the absence of significant uveitis and the presence of serous retinal detachment with shallow anterior chamber, the most likely diagnosis is idiopathic central serous chorioretinopathy.\par
\smallskip\par
\texttt{<answer>}\par
Idiopathic central serous chorioretinopathy\par
\texttt{</answer>}
}
\end{tcolorbox}

\begin{tcolorbox}[breakable, enhanced, colback=red!3, colframe=red!55, boxrule=0.35pt, arc=1.5pt, left=3pt, right=3pt, top=3pt, bottom=3pt, title={\textbf{GPT-5.0 (wrong)}}, fonttitle=\small]
{\scriptsize\raggedright
\textbf{Prediction:} Posterior scleritis\par
\textbf{Judge reason:} Posterior scleritis is a different inflammatory ocular condition and is not a synonym or close paraphrase of central serous chorioretinopathy.\par
\medskip\noindent\textbf{Raw response:}\par
Reasoning:\par
Unilateral serous retinal detachment with mild ocular pain, shallow anterior chamber, axial shortening, and a paradoxical myopic shift indicates anterior displacement from ciliochoroidal effusion, favoring posterior scleritis. Mild vitreous cells and peripapillary/macular detachment support inflammation rather than CSC. Vogt--Koyanagi--Harada is unlikely given the unilateral involvement and lack of significant anterior chamber inflammation or systemic features.\par
\smallskip\par
\texttt{<answer>}\par
Posterior scleritis\par
\texttt{</answer>}
}
\end{tcolorbox}

\begin{tcolorbox}[breakable, enhanced, colback=red!3, colframe=red!55, boxrule=0.35pt, arc=1.5pt, left=3pt, right=3pt, top=3pt, bottom=3pt, title={\textbf{MedReason-8B (wrong)}}, fonttitle=\small]
{\scriptsize\raggedright
\textbf{Prediction:} Persistent fetal vasculature (PFV)\par
\textbf{Judge reason:} Persistent fetal vasculature (PFV) is a different retinal developmental anomaly and is not a synonym or close paraphrase of central serous chorioretinopathy.\par
\medskip\noindent\textbf{Raw response:}\par
Based on the provided clinical details, the patient presents with unilateral visual impairment and blurring of vision in the left eye, along with mild ocular pain. The key findings include:\par
\smallskip\par
1. **Shallow anterior chamber** in the left eye compared to the right.\par
2. **Subretinal fluid** in the left eye, particularly around the optic disc and extending into the macula.\par
3. Normal intraocular pressure bilaterally.\par
4. No anterior chamber inflammation observed during slit-lamp examination.\par
5. A history of being diagnosed with Vogt-Koyanagi-Harada (VKH) disease, which typically involves bilateral involvement and uveitis.\par
\smallskip\par
The presence of a shallow anterior chamber suggests a structural abnormality, possibly related to the anatomy of the eye. However, VKH disease is usually characterized by bilateral symptoms and inflammation, which are not evident here. The absence of anterior chamber inflammation and the unilateral nature of the presentation make VKH less likely.\par
\smallskip\par
Given these observations, the most plausible explanation for the unilateral subretinal fluid and shallow anterior chamber without inflammation could be a congenital anomaly such as persistent fetal vasculature (PFV), also known as persistant hyperplastic primary vitreous (PHPV). This condition can lead to structural abnormalities like a shallow anterior chamber and subretinal fluid due to improper development of the hyaloid artery and lens-related anomalies.\par
\smallskip\par
Therefore, the final diagnosis aligning with the clinical findings would be **persistent fetal vasculature (PFV)**.
}
\end{tcolorbox}

\subsection{Case 2: Behçet's Disease}

\begin{tcolorbox}[breakable, enhanced, colback=blue!3, colframe=blue!55, boxrule=0.35pt, arc=1.5pt, left=3pt, right=3pt, top=3pt, bottom=3pt, title={\textbf{Task Prompt}}, fonttitle=\small]
{\scriptsize\raggedright
Please reason through the following clinical case and predict the final diagnosis. Do not use tables or bullet points. Return your answer in the following format: \texttt{<think>} [Reasoning here] \texttt{</think>} \texttt{<answer>} [Final diagnosis here] \texttt{</answer>}.
}
\end{tcolorbox}

\begin{tcolorbox}[breakable, enhanced, colback=gray!3, colframe=gray!55, boxrule=0.35pt, arc=1.5pt, left=3pt, right=3pt, top=3pt, bottom=3pt, title={\textbf{Full Clinical Case}}, fonttitle=\small]
{\scriptsize\raggedright
An 8-year-old boy of Brazilian descent with autism presented with an 8-month history of painful oral ulcers. He had a one-time perianal ulcer, folliculitis on his scalp and buttocks, intermittent fever, and bloody stools. He had no genital ulcers. He was initially treated as PFAPA with steroids, but symptoms recurred when steroids were tapered. Family history was noncontributory.\par
\smallskip\par
On examination, he had multiple large oral ulcers on the tongue, palate, and lips. Abdominal examination was normal, with no tenderness or masses; there were no perianal ulcers, arthritis, or other skin findings. An extensive workup for systemic lupus erythematosus, herpes simplex virus, Epstein--Barr virus, tuberculosis, cyclic neutropenia, primary immunodeficiencies, and periodic fever syndromes was unremarkable. Ophthalmological examination was negative, with no pathergy. Buccal biopsies showed chronic inflammation with granulation tissue, no granulomas, and grew Fusobacterium periodontium.\par
\smallskip\par
Prometheus IBD sgi testing demonstrated a pattern consistent with Crohn's disease (positive anti-A4-Fa2 IgG, anti-Fax IgG, ATG16L1 heterozygous, NKX2-3 heterozygous, ICAM-1, SAA 179, elevated CRP; ASCA and ANCA negative). Multiple endoscopies and colonoscopies with ileal intubation were negative. Capsule endoscopy, CT of the abdomen, and MRI enterography showed no abnormalities.\par
\smallskip\par
Despite treatment with steroids, colchicine, infliximab, and methotrexate--later escalated to high-dose infliximab--the patient's bloody stools worsened and he developed severe malnutrition (BMI z-score -2.38). At month 18, a fourth colonoscopy revealed large, round ulcers in the terminal ileum with granulation tissue and chronic inflammatory infiltrates, but no granulomas or cobblestoning.\par
\smallskip\par
Treatment was switched to ustekinumab, after which gastrointestinal symptoms improved. At month 19, he developed bilateral hip effusions and later presented with headaches, emesis, and fevers. A lumbar puncture showed an opening pressure of 39 cm H2O; brain MRI/MRA/MRV revealed extensive thrombosis of the superior sagittal sinus, straight sinus, and proximal right transverse sinus. A hypercoagulability panel was negative.
}
\end{tcolorbox}

\noindent\textbf{Model Outputs.}

\begin{tcolorbox}[breakable, enhanced, colback=green!3, colframe=green!55, boxrule=0.35pt, arc=1.5pt, left=3pt, right=3pt, top=3pt, bottom=3pt, title={\textbf{MedGuideX (correct)}}, fonttitle=\small]
{\scriptsize\raggedright
\textbf{Prediction:} Behçet's disease\par
\textbf{Judge reason:} Exact match with the gold diagnosis.\par
\medskip\noindent\textbf{Raw response:}\par
The patient is an 8-year-old boy with a prolonged history of painful oral ulcers, bloody stools, and intermittent fever, initially misdiagnosed as PFAPA. Despite negative workup for common infectious and autoimmune causes, and negative endoscopic and imaging studies for Crohn's disease, he had a positive Prometheus IBD sgi test pattern consistent with Crohn's disease. However, the endoscopic findings were atypical: large round ulcers with granulation tissue and chronic inflammation but no granulomas or cobblestoning, and no ileal involvement on capsule endoscopy, CT, or MRI enterography. His symptoms worsened with multiple Crohn's-directed therapies, including infliximab, and he developed severe malnutrition. The key turning point was the development of bilateral hip effusions and then cerebral venous sinus thrombosis (CVST) with extensive thrombosis of the superior sagittal, straight, and right transverse sinuses, despite a negative hypercoagulability panel. This presentation is highly suggestive of Behçet's disease, which can present with oral ulcers, gastrointestinal involvement (often ileal or colonic ulcers), and thrombotic complications such as CVST. The absence of genital ulcers does not exclude Behçet's disease, as it can be atypical in children. The negative ASCA and ANCA, along with the lack of granulomas, further supports a diagnosis of Behçet's disease rather than Crohn's disease. The response to ustekinumab, which is effective in Behçet's disease, also supports this diagnosis.\par
\smallskip\par
\texttt{<answer>}\par
Behçet's disease\par
\texttt{</answer>}
}
\end{tcolorbox}

\begin{tcolorbox}[breakable, enhanced, colback=red!3, colframe=red!55, boxrule=0.35pt, arc=1.5pt, left=3pt, right=3pt, top=3pt, bottom=3pt, title={\textbf{GPT-5.0 (wrong)}}, fonttitle=\small]
{\scriptsize\raggedright
\textbf{Prediction:} Crohn's disease\par
\textbf{Judge reason:} Crohn's disease is a different inflammatory bowel disease and not a synonym or direct equivalent of Behçet's disease.\par
\medskip\noindent\textbf{Raw response:}\par
Reasoning:\par
Severe, recurrent oral ulcers with bloody stools and later terminal ileal ulcers showing chronic inflammation, plus an IBD serologic pattern, point to Crohn's disease; symptoms improved with ustekinumab. Behçet disease is less likely given absent genital ulcers, negative pathergy and ophthalmologic exams, and GI findings ultimately localized to terminal ileum consistent with Crohn's. Cerebral venous sinus thrombosis is a known IBD complication.\par
\smallskip\par
\texttt{<answer>}\par
Crohn's disease\par
\texttt{</answer>}
}
\end{tcolorbox}

\begin{tcolorbox}[breakable, enhanced, colback=red!3, colframe=red!55, boxrule=0.35pt, arc=1.5pt, left=3pt, right=3pt, top=3pt, bottom=3pt, title={\textbf{MedReason-8B (wrong)}}, fonttitle=\small]
{\scriptsize\raggedright
\textbf{Prediction:} Crohn's disease\par
\textbf{Judge reason:} Crohn's disease is not the same diagnosis as Behçet's disease; they can overlap clinically but are distinct conditions.\par
\medskip\noindent\textbf{Raw response:}\par
Crohn's disease
}
\end{tcolorbox}

\subsection{Case 3: Infective Endocarditis with Renal Embolization}

\begin{tcolorbox}[breakable, enhanced, colback=blue!3, colframe=blue!55, boxrule=0.35pt, arc=1.5pt, left=3pt, right=3pt, top=3pt, bottom=3pt, title={\textbf{Task Prompt}}, fonttitle=\small]
{\scriptsize\raggedright
Please reason through the following clinical case and predict the final diagnosis. Do not use tables or bullet points. Return your answer in the following format: \texttt{<think>} [Reasoning here] \texttt{</think>} \texttt{<answer>} [Final diagnosis here] \texttt{</answer>}.
}
\end{tcolorbox}

\begin{tcolorbox}[breakable, enhanced, colback=gray!3, colframe=gray!55, boxrule=0.35pt, arc=1.5pt, left=3pt, right=3pt, top=3pt, bottom=3pt, title={\textbf{Full Clinical Case}}, fonttitle=\small]
{\scriptsize\raggedright
A 52-year-old woman with known unrepaired Type II tricuspid atresia with d-transposition of the great arteries, hypoplastic right ventricle, unrestricted atrial and ventricular septal defects, and multi-level pulmonic stenosis presented with sudden-onset severe right upper quadrant abdominal pain without nausea, vomiting, diarrhea, or dysuria. Two months earlier, she had flu-like symptoms treated with doxycycline, and she reported poor dentition with no dental care for two years. On examination, her blood pressure was 90/54 mmHg, she was afebrile, heart rate was 80 bpm, and oxygen saturation was 79\% on room air. A grade-2 systolic ejection murmur was heard at the left upper sternal border. She had cyanotic fingers bilaterally but no Osler nodes, Janeway lesions, or splinter hemorrhages. Laboratory studies showed hemoglobin 17.9 g/dL and C-reactive protein 2.2 mg/dL; lactate, white blood cell count, and creatinine were within reference ranges. Computed tomography of the abdomen demonstrated an acute right renal infarct; cerebral computed tomography was negative. She was anticoagulated with heparin.
}
\end{tcolorbox}

\noindent\textbf{Model Outputs.}

\begin{tcolorbox}[breakable, enhanced, colback=green!3, colframe=green!55, boxrule=0.35pt, arc=1.5pt, left=3pt, right=3pt, top=3pt, bottom=3pt, title={\textbf{MedGuideX (correct)}}, fonttitle=\small]
{\scriptsize\raggedright
\textbf{Prediction:} Infective endocarditis with septic embolization to the right kidney\par
\textbf{Judge reason:} The prediction includes the gold diagnosis, infective endocarditis; the added septic embolization to the right kidney is a complication and does not change the core diagnosis.\par
\medskip\noindent\textbf{Raw response:}\par
The patient has a complex cyanotic congenital heart disease with unrepaired tricuspid atresia, d-transposition of the great arteries, hypoplastic right ventricle, unrestricted atrial and ventricular septal defects, and multi-level pulmonic stenosis. This anatomy results in right-to-left shunting and systemic cyanosis, as evidenced by her low oxygen saturation and cyanotic fingers. She presents with sudden severe right upper quadrant abdominal pain, and imaging reveals an acute right renal infarct. Her history includes recent doxycycline use for flu-like symptoms and poor dentition, which raises concern for a source of bacteremia. Although she lacks classic peripheral stigmata of infective endocarditis such as Osler nodes or Janeway lesions, the presence of a new renal infarct in the setting of cyanotic heart disease is highly suggestive of septic emboli from infective endocarditis. The murmur at the left upper sternal border may represent her baseline murmur from pulmonic stenosis, but the acute presentation with organ infarction points to embolic complications. Given her cyanotic heart disease, she is at high risk for infective endocarditis due to turbulent flow and endothelial damage, and the renal infarct is a classic manifestation of septic emboli. Therefore, the most likely diagnosis is infective endocarditis with septic embolization to the kidney.\par
\smallskip\par
\texttt{<answer>}\par
Infective endocarditis with septic embolization to the right kidney\par
\texttt{</answer>}
}
\end{tcolorbox}

\begin{tcolorbox}[breakable, enhanced, colback=red!3, colframe=red!55, boxrule=0.35pt, arc=1.5pt, left=3pt, right=3pt, top=3pt, bottom=3pt, title={\textbf{GPT-5.0 (wrong)}}, fonttitle=\small]
{\scriptsize\raggedright
\textbf{Prediction:} Paradoxical systemic arterial embolism causing acute right renal infarction in cyanotic congenital heart disease (unrepaired tricuspid atresia with d-TGA)\par
\textbf{Judge reason:} Predicted diagnosis is paradoxical systemic arterial embolism/renal infarction due to cyanotic congenital heart disease, which is not the same as infective endocarditis.\par
\medskip\noindent\textbf{Raw response:}\par
Reasoning:\par
Unrepaired cyanotic congenital heart disease with right-to-left shunt allows venous thrombi to bypass the lungs and embolize systemically. She is afebrile with normal WBC, only mildly elevated CRP, no peripheral stigmata of endocarditis, and acute CT-proven renal infarct; anticoagulation was initiated for embolism. Infective endocarditis with septic emboli is less likely given absence of fever and classic signs.\par
\smallskip\par
\texttt{<answer>}\par
Paradoxical systemic arterial embolism causing acute right renal infarction in cyanotic congenital heart disease (unrepaired tricuspid atresia with d-TGA)\par
\texttt{</answer>}
}
\end{tcolorbox}

\begin{tcolorbox}[breakable, enhanced, colback=red!3, colframe=red!55, boxrule=0.35pt, arc=1.5pt, left=3pt, right=3pt, top=3pt, bottom=3pt, title={\textbf{MedReason-8B (wrong)}}, fonttitle=\small]
{\scriptsize\raggedright
\textbf{Prediction:} Intracardiac thrombus\par
\textbf{Judge reason:} Predicted diagnosis is intracardiac thrombus with paradoxical embolism/renal infarction, which is not the same as infective endocarditis.\par
\medskip\noindent\textbf{Raw response:}\par
\#\#\# Finding Reasoning Paths:\par
1. **Right upper quadrant abdominal pain**: Consider causes such as gallbladder disease (e.g., cholecystitis), pancreatitis, peptic ulcer disease, or vascular events like mesenteric ischemia or renal infarction.\par
2. **History of flu-like illness and doxycycline use**: Evaluate for complications related to recent antibiotic use (e.g., drug-induced side effects) or secondary infections.\par
3. **Cyanosis and low oxygen saturation**: Investigate cardiac or pulmonary causes, including pulmonary embolism, chronic lung disease, or congenital heart defects.\par
4. **Acute renal infarction on CT scan**: Explore potential causes such as thromboembolism, vasculitis, or other vascular abnormalities.\par
5. **Congenital heart disease history**: Assess how underlying structural heart defects (e.g., tricuspid atresia, transposition of the great arteries) contribute to systemic issues like hypoxemia or embolic events.\par
\smallskip\par
---\par
\smallskip\par
\#\#\# Reasoning Process:\par
1. **Initial Presentation**:\par
   - The patient has severe right upper quadrant abdominal pain, cyanosis, and hypoxemia (oxygen saturation 79\%). These findings suggest a systemic issue involving both cardiovascular and possibly gastrointestinal systems.\par
   - The absence of fever, nausea, vomiting, or diarrhea makes infectious or inflammatory causes less likely in the GI tract.\par
\smallskip\par
2. **Cardiac History**:\par
   - The patient has a complex congenital heart defect (tricuspid atresia with transposition of the great arteries). This condition predisposes her to cyanosis due to impaired oxygenation and potential complications like paradoxical emboli.\par
   - Cyanosis and hypoxemia could also result from pulmonary hypertension or right-to-left shunting due to the congenital defect.\par
\smallskip\par
3. **Renal Infarction**:\par
   - The CT scan shows an acute right renal infarction, which is often caused by embolism, thrombosis, or vasculitis. In this case, embolism is most likely given the systemic nature of the presentation.\par
   - Emboli can originate from the heart, particularly in patients with congenital heart disease, where intracardiac thrombi may form due to stagnant blood flow or turbulent currents.\par
\smallskip\par
4. **Doxycycline Use**:\par
   - Doxycycline is not commonly associated with embolic phenomena, so it is unlikely to be directly responsible for the renal infarction.\par
   - However, the flu-like illness treated with doxycycline raises the possibility of a preceding viral infection that could have triggered a hypercoagulable state or endothelial damage, increasing the risk of thromboembolism.\par
\smallskip\par
5. **Systemic Embolism**:\par
   - The combination of cyanosis, hypoxemia, and renal infarction strongly suggests a systemic embolic event. Given the patient's congenital heart disease, paradoxical embolism (via a right-to-left shunt) is a plausible mechanism.\par
   - Intracardiac thrombus formation is a common complication in patients with congenital heart defects, especially when there are atrial or ventricular septal defects allowing shunting of deoxygenated blood.\par
\smallskip\par
6. **Exclusion of Other Causes**:\par
   - There is no evidence of infection (no fever, leukocytosis, or elevated CRP).\par
   - No signs of pancreatitis or peptic ulcer disease based on the clinical presentation.\par
   - The absence of neurological symptoms makes stroke or cerebral embolism less likely.\par
\smallskip\par
---\par
\smallskip\par
\#\#\# Conclusion:\par
The most likely diagnosis is **intracardiac thrombus formation**, leading to a paradoxical embolism causing the acute right renal infarction and systemic hypoxemia. This conclusion aligns with the patient's congenital heart disease, cyanosis, and renal infarction findings.\par
\smallskip\par
Final Answer\par
\smallskip\par
Intracardiac thrombus
}
\end{tcolorbox}

\section{AI Use Statement}
\label{sec:ai_use}
We used generative AI tools to polish the manuscript for clarity and readability and to assist in finding relevant prior work. GPT-5.4 with xhigh reasoning performed recommendation extraction, decision-tree generation, and decision-tree validation. A deterministic converter generated the Python functions. GPT-5.4 with xhigh reasoning also generated synthetic factual and counterfactual question text and reasoning traces. GPT-6.1 Sol with xhigh reasoning generated the held-out guideline cases and their action pseudolabels; these pseudolabels were used directly in the action evaluation. The data generation and validation procedures are detailed in Section~\ref{sec:data} and Appendix~\ref{sec:data_details}. The authors reviewed the AI-assisted text and take responsibility for the final content of this work.

\clearpage
\section{Prompts for the Held-Out Guideline Evaluation}
\label{sec:heldout_prompts}
The first two instructions summarize the generation and student tasks in Appendix~\ref{sec:heldout_eval_protocol}. The third gives the written-path judge template in full.

\paragraph{Case generation.}
Given the guideline passage, construct an original case with the patient variables needed to determine the recommended action. Make two variants, each changing one variable from the original. In one variant the guideline still recommends the original action; in the other it recommends a different action. Supply the reference action for each version.

\paragraph{Model response.}
Given the guideline passage, one case and its question, state the recommended action and the patient condition that determines it. Present the three versions in separate queries.

\paragraph{Written-path judge.}
\begingroup\small\raggedright
\emph{Evaluate whether the response gives a correct guideline-based decision path for this patient.}\par\smallskip
Guideline passage\newline
\texttt{\{guideline\_passage\}}\par
Patient case and question\newline
\texttt{\{case\_and\_question\}}\par
Reference path prepared from the guideline\newline
\texttt{\{decisive\_conditions\_applicable\_exceptions\_and\_accepted\_actions\}}\par
Anonymous response\newline
\texttt{\{model\_response\}}\par\smallskip
Use the supplied guideline as the authority. Grade only the reasoning stated in the response. Award 1 only if all three criteria are met:
\begin{enumerate}[nosep,leftmargin=*]
\item The response identifies the patient fact or facts that determine the action.
\item It correctly applies the relevant guideline rule to those facts, including any threshold or exception that determines which branch applies.
\item Its recommended action follows from that branch and answers the question.
\end{enumerate}
\smallskip
Award 0 if the action is correct but the response omits the decisive reason, uses the wrong branch, or contradicts a decisive patient fact. Accept concise reasoning and any alternative path supported by the passage. Do not require the reference wording, an explicit rule citation, or a list of irrelevant branches. Do not supply missing reasoning on the respondent's behalf.\par\smallskip
If the passage and case do not determine which actions are acceptable, or if the reference conflicts with the passage, return \texttt{"review"} with a null score.\par\smallskip
Return JSON only:
{\fontsize{8}{10}\selectfont
\begin{verbatim}
{
  "verdict": "pass | fail | review",
  "score": "1, 0, or null as a JSON number or null",
  "guideline_evidence": "Shortest exact passage supporting the judgment",
  "response_evidence": "Exact phrase applying the decisive condition, or null",
  "reason": "One sentence explaining the judgment"
}
\end{verbatim}
}
\endgroup

%% file: tables/paired_analysis.tex
\begin{table}[t]
\caption{Paired answer transition analysis between Qwen3.5 and \model. Corrected denotes cases where \model changes a wrong Qwen3.5 answer into a correct one, while Regressed denotes the opposite transition. Unchanged denotes cases with no change in correctness status. Net Gain is computed as Corrected minus Regressed.}
\label{tab:transition}
\centering
\scriptsize
\setlength{\tabcolsep}{2.5pt}
\renewcommand{\arraystretch}{1.15}
\resizebox{\columnwidth}{!}{
\begin{tabular}{@{}lcccccccc@{}}
\toprule
\multicolumn{1}{c}{\textbf{Scale}} & \multicolumn{4}{c}{\textbf{4B}} & \multicolumn{4}{c}{\textbf{9B}} \\
\cmidrule(lr){2-5}\cmidrule(l){6-9}
\multicolumn{1}{c}{\textbf{Benchmark}}
& \textbf{Corrected} & \textbf{Regressed} & \textbf{Unchanged} & \textbf{Net Gain}
& \textbf{Corrected} & \textbf{Regressed} & \textbf{Unchanged} & \textbf{Net Gain} \\
\midrule
MedCaseReasoning & 87 & 27 & 783 & \textcolor{langgreen}{\textbf{+60}}
                 & 133 & 55 & 709 & \textcolor{langgreen}{\textbf{+78}} \\
MedQA            & 76 & 17 & 1,180 & \textcolor{langgreen}{\textbf{+59}}
                 & 110 & 22 & 1,141 & \textcolor{langgreen}{\textbf{+88}} \\
MIMIC-CDM-FI     & 20 & 0 & 980 & \textcolor{langgreen}{\textbf{+20}}
                 & 42 & 6 & 952 & \textcolor{langgreen}{\textbf{+36}} \\
ER-Reason        & 23 & 0 & 977 & \textcolor{langgreen}{\textbf{+23}}
                 & 40 & 10 & 950 & \textcolor{langgreen}{\textbf{+30}} \\
\bottomrule
\end{tabular}
}
\end{table}

%% file: tables/guideline_baselines.tex
\begin{table*}[t]
\caption{Accuracy of guideline-based methods on four clinical reasoning benchmarks in percent. Figure~\ref{fig:ablation}b shows the corresponding gains over each Qwen3.5 backbone in percentage points. Bold marks the best score at each scale.}
\label{tab:guideline_baselines}
\centering
\fontsize{9}{10}\selectfont
\setlength{\tabcolsep}{1.5pt}
\renewcommand{\arraystretch}{1.2}
\begin{tabular}{@{}L{0.44\textwidth}*{5}{C{\dimexpr0.112\textwidth-2\tabcolsep\relax}}@{}}
\toprule
\multicolumn{1}{c}{\textbf{Model and method}}
& \makecell[c]{\textbf{MedCase}\\\textbf{Reasoning}}
& \makecell[c]{\textbf{MedQA}}
& \makecell[c]{\textbf{MIMIC-}\\\textbf{CDM-FI}}
& \makecell[c]{\textbf{ER-}\\\textbf{Reason}}
& \textbf{Average} \\
\midrule
\rowcolor{gray!10}Qwen3.5-4B & 28.65 & 72.43 & 80.40 & 23.60 & 51.27 \\
+ RAG with Guidelines & 28.54 & 73.45 & 81.90 & \textbf{26.30} & 52.55   \\
+ 3-Shot In-Context Learning & 28.65 & 71.41 & 81.40 & 26.10 & 51.89 \\
\rowcolor{figurelightblue}\textbf{\model-4B} & \textbf{35.34} & \textbf{77.06} & \textbf{82.40} & 25.90 & \textbf{55.18} \\
\midrule
\rowcolor{gray!10}Qwen3.5-9B & 32.66 & 73.21 & 81.60 & 25.70 & 53.29 \\
+ RAG with Guidelines & 31.88 & 74.71 & 82.50 & 27.40 & 54.12 \\
+ 3-Shot In-Context Learning & 33.11  & 74.94 & 80.10 & 28.20 & 54.09 \\
+ CPGPrompt~\citep{deng2026cpgprompt} & 14.16  & 59.15 & 67.60 & 19.90 & 40.20 \\
+ Fine-tuning with CPG & 28.87 & 65.04 & 78.40 & 21.50 & 48.45 \\
+ RL with CPG-Derived Process Rewards & 37.68 & 75.73 & 83.40 & 27.80 & 56.15 \\
\rowcolor{figurelightblue}\textbf{\model-9B} & \textbf{41.36} & \textbf{80.13} & \textbf{85.20} & \textbf{28.70} & \textbf{58.85} \\
\bottomrule
\end{tabular}
\end{table*}